\documentclass[journal]{IEEEtran}

\ifCLASSINFOpdf
\else
\fi
\usepackage[utf8]{inputenc} 
\usepackage[T1]{fontenc}    
\usepackage{url}            
\usepackage{booktabs}       
\usepackage{amsfonts}       
\usepackage{nicefrac}       
\usepackage{microtype}      
\usepackage{bm}
\usepackage{multirow}
\usepackage{array}
\usepackage{threeparttable}
\usepackage{graphicx}
\usepackage{color}
\usepackage{wrapfig}
\usepackage{caption2}
\usepackage{amsmath}
\usepackage{algpseudocode} 
\usepackage{algorithmicx,algorithm}
\usepackage{calc}
\usepackage{linegoal}
\usepackage{graphicx}
\usepackage{subfigure}

\usepackage{url}
\usepackage[colorlinks,
            linkcolor=red,
            anchorcolor=blue,
            citecolor=green
            ]{hyperref}

\begin{document}
%
\title{Unsupervised Multi-view Clustering by Squeezing Hybrid Knowledge from Cross View and Each View}
%
%
%

\author{Junpeng Tan,  Yukai Shi, Zhijing Yang, Caizhen Wen, Liang Lin

}

\markboth{}%
{Shell \MakeLowercase{\textit{et al.}}: Bare Demo of IEEEtran.cls for IEEE Journals}
%



\maketitle

\begin{abstract}

Multi-view clustering methods have been a focus in recent years because of their superiority in clustering performance. However, typical traditional multi-view clustering algorithms still have shortcomings in some aspects, such as removal of redundant information, utilization of various views and fusion of multi-view features. In view of these problems, this paper proposes a new multi-view clustering method, low-rank subspace multi-view clustering based on adaptive graph regularization. We construct two new data matrix decomposition models into a unified optimization model. In this framework, we address the significance of the common knowledge shared by the cross view and the unique knowledge of each view by presenting new low-rank and sparse constraints on the sparse subspace matrix. To ensure that we achieve effective sparse representation and clustering performance on the original data matrix, adaptive graph regularization and unsupervised clustering constraints are also incorporated in the proposed model to preserve the internal structural features of the data. Finally, the proposed method is compared with several state-of-the-art algorithms. Experimental results for five widely used multi-view benchmarks show that our proposed algorithm surpasses other state-of-the-art methods by a clear margin.
\end{abstract}


%
\IEEEpeerreviewmaketitle

\begin{figure*}[t]
  \centering
  \includegraphics[width=0.8\textwidth]{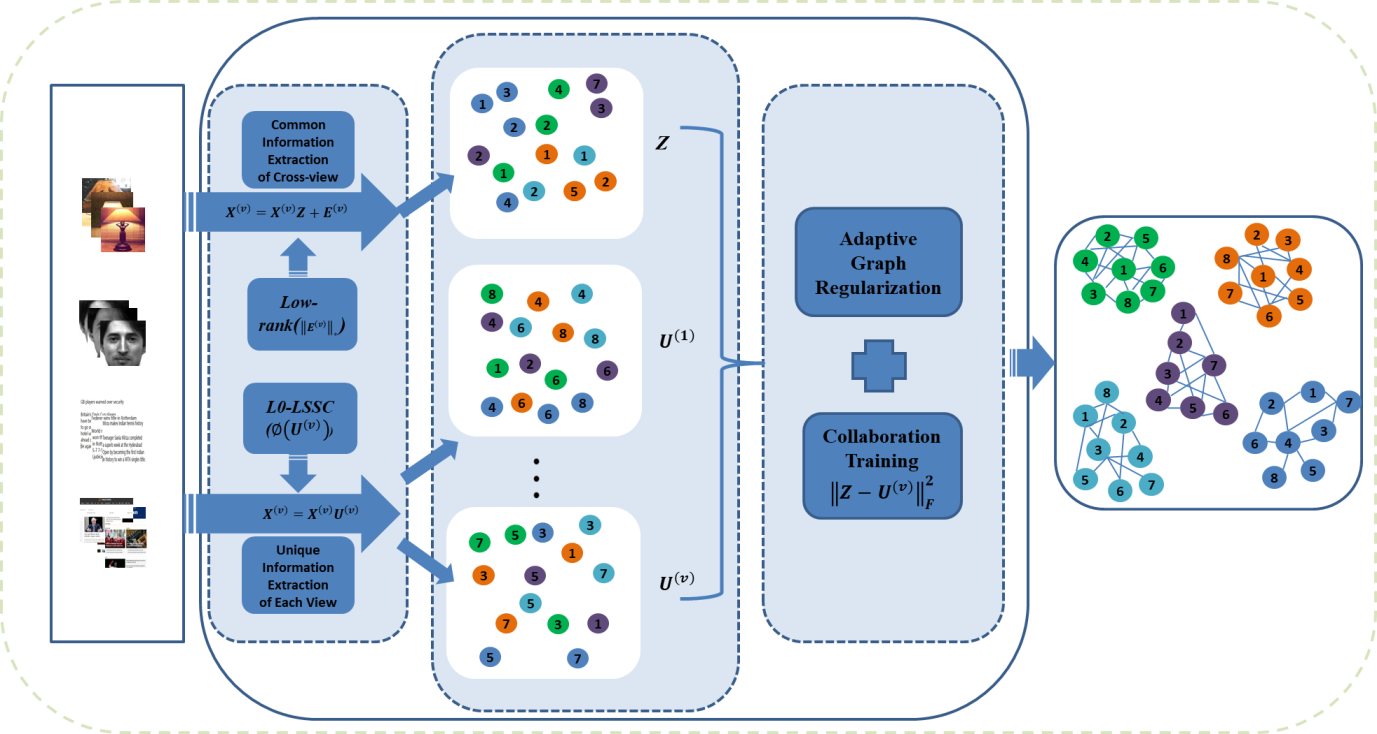}
  \caption{The flow chart of the proposed UMC-CEV algorithm.} 
  \label{fig:UMC-CEV algorithm}
\end{figure*}

\section{Introduction}
Multi-view clustering methods are a research subject that has received much attention in recent years. The definition of multi-view clustering is potential to have very broad in the field of pattern recognition~\cite{zhang2019scan,zhao2017deeply}, computer vision~\cite{cao2017attention,shi2020ddet,shi2017structure} and machine intelligence~\cite{lu2020end,shi2019face}. Some researchers define it as an object to extract its information from different angles and then combine related information from each angle. In addition, some researchers define multi-view as multi-modal \cite{[1]wu2013sparse}, which improves the performance of clusters based on different modal information, such as images, text and speech. Furthermore, multi-view can also be understood as a combination of multiple features \cite{[2]journals/taffco/ChenCCF18}. We can use a classic method to represent the characteristics of an image or text, such as SIFT \cite{[3]journals/ijcv/Lowe04}, LBP \cite{[4]journals/pami/AhonenHP06}, and GABOR \cite{[5]journals/pami/Lee96a}. In this paper, multi-view is regarded as combining multiple features to improve the performance of clusters. With the development of multi-view clustering, relevant researchers have proposed many clustering algorithms for multi-view feature combination. We can divide multi-view clustering algorithms according to different levels. For the hierarchical level of clustering, we can classify multi-view clustering algorithms according to the use of training and the number of samples as supervised multi-view clustering \cite{[6]journals/nca/WangLYM17}, semi-supervised multi-view clustering \cite{[7]conf/ijcai/ZhangLLHLZ18} and unsupervised multi-view clustering \cite{[8]journals/tkde/HouNTY17}. The earliest algorithm was the subspace-based supervised clustering algorithms \cite{[9]conf/cvpr/TulsianiZEM17,[10]conf/cvpr/XuHN16}, which require the label data of all samples to be known and need to manually select the training samples. Although this method can promise state-of-the-art clustering performance, it consumes large amounts of manual effort and time to a certain extent. In addition, semi-supervised multi-view algorithms based on low-rank sparse total spatial clustering have also appeared \cite{[11]journals/tip/NieCLL18,[12]conf/aaai/NieCL17,[13]journals/spl/XueDDLHL19} in recent years. Although these methods reduce the sample labels and training samples that need to be labelled compared to the method of supervised learning, some defects still exist. For example, if we do not completely know the sample clustering labels, we cannot use this method to effectively cluster the data. Given this drawback, an unsupervised clustering method is needed. 

	Since it does not require training samples in advance for single-view \cite{[14]lu2018subspace,[15]journals/tsp/LiV16,[16]journals/nn/ZhanWHWF19} or multi-view tasks \cite{[17]yang2019skeletonnet,[18]ma2018learning,[19]zhang2015object,li2020multi}, unsupervised clustering has been the focus of research in recent years and the difficulty of research. Especially in the field of multi-view clustering algorithms, through the continuous efforts of relevant scientific researchers, many unsupervised clustering algorithms have been proposed. Xu et al. \cite{[20]xu2015multi} proposed an unsupervised multi-view intact spatial learning (MISL) algorithm that integrated encoded supplementary information into multiple views to discover a potential complete representation of the data. From the perspective of multi-view feature combination, we can use roughly three categories of multi-view methods: subspace clustering \cite{[21]conf/iccv/GaoNLH15}, low-rank sparse representation \cite{[22]journals/isci/ZhangSLRCL19} and adaptive graph learning \cite{[23]wen2018incomplete}. These works showed that complementary information between views is beneficial for the classification performance. To make full use of the diversity of views, Cao et al. \cite{[24]cao2015diversity} proposed diversity-induced multi-view subspace clustering (DiMSC) using the Hilbert-Schmidt independence criterion (HSIC) to obtain the complementary information of each view. In addition, Wang et al. \cite{[25]wang2017exclusivity} used a low-rank strategy to extract the complementary information from each view by the $L_1$-norm, called exclusivity-consistency regularization. These methods considered the use of the complementary information between views but did not consider the common information sharing by each view. In view of this, Ding et al. \cite{[26]ding2017robust} proposed robust multi-view data analysis through a collective low-rank subspace (CLRS). This method used view information to extract the common information existing in each view. With the development of algorithms in recent years, some ways to extract common information between views have been proposed \cite{[27]kumar2011co,[28]zhan2018graph,[29]zhan2018multiview}. The simplest method is to perform a weighted fusion among multiple views. Zhao et al. \cite{[30]zhao2017consensus} proposed an adaptive weighted fusion algorithm for each view. Another kind of methods extracts common information between views not only through the constraints of the model but also using data decomposition of sparse representation for the original data. This idea can better reflect the correlation between the feature information of each view. Recently, Wang et al. \cite{[31]wang2019multi} proposed a new multi-view fusion subspace clustering algorithm (MSC-IAS) that combined the information of each view by adaptively decomposing the original data and used an adaptive manifold to preserve the data structure. The common information and complementary information between views are both important for the representation of an object. Recently, relevant researchers have proposed a new idea for multi-view information utilization that decomposes the original data into the common information and complementary information of each view. This approach can extract the multi-view features to a certain extent and preserve the structural features of the original data when it is decomposed. Luo et al. \cite{[32]luo2018consistent} proposed a novel subspace multi-view clustering algorithm called consistent and specific multi-view subspace clustering (CSMSC). The main idea of this method was to deal with a sparse representation matrix after the decomposition of a multi-view original matrix. The sparse representation matrix for each view was represented by a common feature matrix and each view-specific feature matrix. However, there are still some drawbacks in practice. When dealing with a sparse representation matrix, this method will cause information loss for each view and structural connection between views. To solve this problem, Tang et al. \cite{[33]tang2019cross} proposed cross-view local structure preserved diversity and consensus learning for multi-view unsupervised feature selection (CPV-DCL). In CPV-DCL, the original multi-view data matrix was adaptively weighted, and then the selected multi-views were directly decomposed into a common feature part and distinctive feature part unique to each view. Finally, these two feature parts were subjected to low-rank sparse constraints, and manifold learning was used to preserve local structural features. Although the above-mentioned algorithms can achieve good results when solving a single problem to a certain extent, there are still some shortcomings in dealing with some comprehensive problems. We need to consider the adaptability of the algorithm in many aspects, such as the application scenarios of low-rank sparse constraints, the extraction of common features, the identification of the unique information of each view, and the similarity measure between samples. 
	
	In summary, several issues with clustering multi-view data capture our attention: (1) Remove redundant information from the original matrix, and extract useful information from the error matrix. (2) Make full use of the common information between views and the complementary information of each view. (3) Sparse representation will destroy the feature association between the original samples. Adaptive factor learning is used to maintain the local structural features. (4) Use novel low-rank sparse constraints to solve the existing drawbacks of the nuclear norm and $L_1$-norm.

	To solve the above-mentioned problems, we propose a new multi-view unsupervised clustering method called unsupervised low-rank clustering. The proposed method is a unified framework for unsupervised multi-view clustering that incorporates the common information of the cross-view and the unique information of each view (UMC-CEV). We combine low-rank sparse decomposition with geometric structure retention to extract the common matrix and complementary information of each view. The main idea is to perform two sparse subspace decompositions on the original multi-view data matrix, one of which decomposes the original multi-view data matrix into a common sparse subspace matrix and the error of each view. The other decomposes the original multi-view data matrix into multiple sparse subspace matrices that represent the complementary information between various views, also called diversity. Insufficient data constraints based on the nuclear norm and $L_1$-norm will lead to over-penalized and only approximate original problems. We use a low-rank sparse constraint similar to the nuclear norm and $L_1$-norm to address this problem. In addition, we can use the sparse subspace obtained by decomposing the original data as the similarity measure matrix between different samples and use the sparse matrices for adaptive factor learning. To a certain extent, this approach ensures that the original data are not corrupted when performing sparse representation. Additionally, the discriminant information of intra-classes and inter-classes can be reflected. The specific flow of the proposed algorithm is shown in Fig. 1. The main contributions of this paper can be summarized as follows:
	
	(1) Two kinds of sparse subspace decomposition models are used to process the original multi-view data to extract the common information of cross-view features and the unique information of each view feature.
	
	(2) A new threshold function and singular value decomposition (SVD) are used to replace the low-rank sparse effect of the nuclear norm and $L_1$-norm to solve the over-penalized problem of the nuclear norm and $L_1$-norm. Moreover, the error matrix of each view feature and the common features is subject to the nuclear norm to minimize the error matrix of each view. 
	
	(3) Using the idea of adaptive factor learning, a common dilution subspace matrix is taken as the similarity measure matrix between samples, and the local geometric features are preserved when the original data are decomposed by sparse representation. In addition, the optimal similarity matrix is used for clustering discrimination.
	
	(4) We use a common view feature extraction model, i.e., the sparse representation of each view feature model, the novel low-rank and sparse representation constraints, and adaptive manifold learning as a unified objective function. Finally, we achieve good results for several public datasets by using spectral clustering.
	
	The rest of this paper is organized as follows. Section \uppercase\expandafter{\romannumeral2} briefly introduces the related works. The proposed algorithm is explained in detail in Section \uppercase\expandafter{\romannumeral3}. Section \uppercase\expandafter{\romannumeral4} presents the optimization part of our proposed algorithm. The experimental results and analysis are given in Section \uppercase\expandafter{\romannumeral5}. Section \uppercase\expandafter{\romannumeral6} summarizes the conclusions of this paper.

\section{Related Work}

In this section, we briefly introduce the related technologies involved in our algorithm, including sparse subspace clustering (SSC) \cite{[34]elhamifar2013sparse} and adaptive graph regularization (AGR) \cite{[35]lu2016low}, which are widely used in machine learning and pattern recognition fields. Our proposed multi-view clustering algorithm is based on the improvement of these two algorithms, which will be described in detail in Section \uppercase\expandafter{\romannumeral3}.
\subsection{Sparse Subspace Clustering (SSC)}
SSC has been developed for many years and has received extensive attention in the field of feature extraction and clustering. Recently, some improved methods of SSC have been presented, such as sparse low-rank and sparse constraints, to make SSC more robust. In general, the nuclear norm and $L_1$-norm are used to control the rank and sparsity of a sparse subspace matrix. However, Brbic et al. \cite{[36]brbic2018} found that the nuclear norm and $L_1$-norm over-penalize and just approximate the original problem. Therefore, Brbic et al. \cite{[36]brbic2018} proposed a novel low-rank SSC algorithm called $L_0$-motivated low-rank sparse subspace clustering (L0-LSSC). This algorithm uses a multivariate generalization of minimize-concave penalty (GMC-LRSSC) as a regularization constraint. Its expression is as follows:
\begin{equation}
\begin{aligned}
&\min _{U} \frac{1}{2}\|X-X U\|_{F}^{2}+\lambda \varphi_{B}(\delta(U))+\tau \varphi_{B}(U),\\
&\text {s.t.} \operatorname{diag}(U)=0
\end{aligned}
\end{equation}

\noindent where $\delta\left(U\right)$ denotes the vector containing the singular values of the sparse subspace matrix $U$ and $\varphi_B\left(u\right)$:$\ R^n\rightarrow R$ is the GMC-LRSSC penalty, which can be defined as:

\begin{subequations}
\begin{equation}
\varphi_{B}(u) =\|u\|_{1}-S_{B}(u); \\
\end{equation}
\begin{equation}
S_{B}(u) =\inf _{v \in R^{n}}\left\{\|v\|_{1}+\frac{1}{2}\|B(u-v)\|_{2}^{2}\right\};
\end{equation}
\indent Lemma 1 (\cite{[37]selesnick2017sparse}): Let$ A\in R^{m \times n}$,$u\in R^n$,$y\in R^m$ and $\theta>0$. We can define the function $F(u)\colon R^n\rightarrow R$ as:
\begin{equation}
 F(u)=\frac{1}{2}\|y-Au\|_{2}^{2}+\theta \varphi_{B}(u),
\end{equation}
\end{subequations}

\noindent where $S_{B} \left(U \right)$:$\ R^n\rightarrow R$ denotes the generalized Huber function.     According to Ref. \cite{[37]selesnick2017sparse}, the variable $B$ satisfies $B=\sqrt{\gamma / {\theta A}} (0 \leq \gamma \leq 1)$ in Eq. (2b), if $A^{T} A-\theta B^{T}B$ satisfies positive semi-definite matrix and $F(u)$ is convex function. We can see that the second term of the penalty constraint in Eq. (1) is very similar to the previous nuclear norm, and the third term of the penalty constraint in Eq. (1) is very similar to the previous $L_1$-norm. These novel penalty items can be solved by the corresponding relaxation function. Details of GMC-LRSSC can be found in Refs. \cite{[36]brbic2018,[37]selesnick2017sparse}.

\subsection{Adaptive Graph Regularization (AGR)}

Graph regularization is a useful tool for matrix factorization. Its main function is to preserve the local geometry. However, most similarity graph matrix methods are not adaptive \cite{[38]tang2018learning}. AGR has made a large breakthrough in recent years. In AGR, the similarity matrix continually optimizes the internal geometry of the data during iterations. This guarantees an optimal measure of the similarity between different samples to a certain extent. Zhan et al. \cite{[39]zhan2017graph} proposed a graph learning-based method that improved the quality of the graph. This method used the eigenvalue decomposition method to preserve the local geometry of the similarity matrix. In addition, Wen et al. \cite{[40]wen2018low} proposed a novel AGR method that focused on learning a more general graph. This method used two parts of local geometry, including the original data and the clustering. Furthermore, the method used a low-rank constraint on the similarity matrix and the error matrix. The loss function of this algorithm is as follows:
\begin{equation}
\begin{split}
\min _{Z, E, F} \sum_{i, j}^{n}\left\|x_{i}-x_{j}\right\|_{2}^{2} z_{i j}+\lambda_{1}\|Z\|_{*}\\
+\lambda_{2}\|E\|_{1}+2 \lambda_{3} \operatorname{tr}\left(F^{T} L_{z} F\right), \\
\text { s.t. } X=X Z+E, \operatorname{diag}(Z)=0, \\Z \geq 0, \Sigma_{j} z_{i j}=1, F^{T} F=I
\end{split}
\end{equation}

\noindent where $\lVert . \rVert_*$  is the nuclear constraint on the similarity matrix and can be denoted as  $\lVert Z \rVert_*=\sum_{i}^{n}\delta_i$, where $\delta_i$ is the singular value of the similarity matrix and can reflect the low-rank constraint effect of the matrix. The error matrix constraint can be calculated as $\lVert E \rVert_1=\sum_{i,j=1}^{n}\left|e_{ij}\right|$, which is the sparsity selection constraint. In this loss function, the first term preserves the local geometry of the original data, and variable Z is the similarity matrix. The second and third terms are the similarity matrix low-rank constraint and the error matrix sparsity constraint, respectively. The last term preserves the geometry characteristics of the Laplacian matrix $L_z$ by the eigenvalue decomposition. The parameters $\lambda_1$, $\lambda_2$ and $\lambda_3$ are used to trade off the importance of each constraint. More details of low-rank representation based on adaptive graph regularization (LRR-AGR) can be found in Ref. \cite{[40]wen2018low}. 

\indent As we know, optimization algorithms have always been the focus of machine learning and pattern recognition. Many advanced optimization algorithms have been proposed. For example, Yuan et al. \cite{[41]yuan2020prp} proposed a new conjugate gradient algorithm (PRP-WWP). Li et al.  \cite{[42]li2019dividing} proposed a block-based multi-objective algorithm for the optimization of large-scale feature selection (DMEA-FS).Yong et al. \cite{[43]yong2019novel} proposed a new bat optimization algorithm (BABLUE) based on cross boundary learning (CBL) and uniform exploitation strategy (UES). These three optimization algorithms take the advantages of fast convergence speed, relatively low computational complexity and global optimization. But they are still inadequate for dealing with non-convex and non-smooth models. Besides, these three optimization algorithms will introduce more optimization variables during the optimization. According to the research of relevant researchers, the alternating direction method of multipliers (ADMM) is an effective optimization method to solve non-convex, non-smooth and non-Lipschitzand models. It can not only process a large amount of data and introduce a small number of optimization variables, but also adapt the Lagrange multiplier introduced by optimization according to the characteristics of the model. In view of the model features and advantages of ADMM, we use it to solve the proposed model.

\section{The Proposed Model}
\begin{figure}
  \centering
  \includegraphics[width=0.48\textwidth]{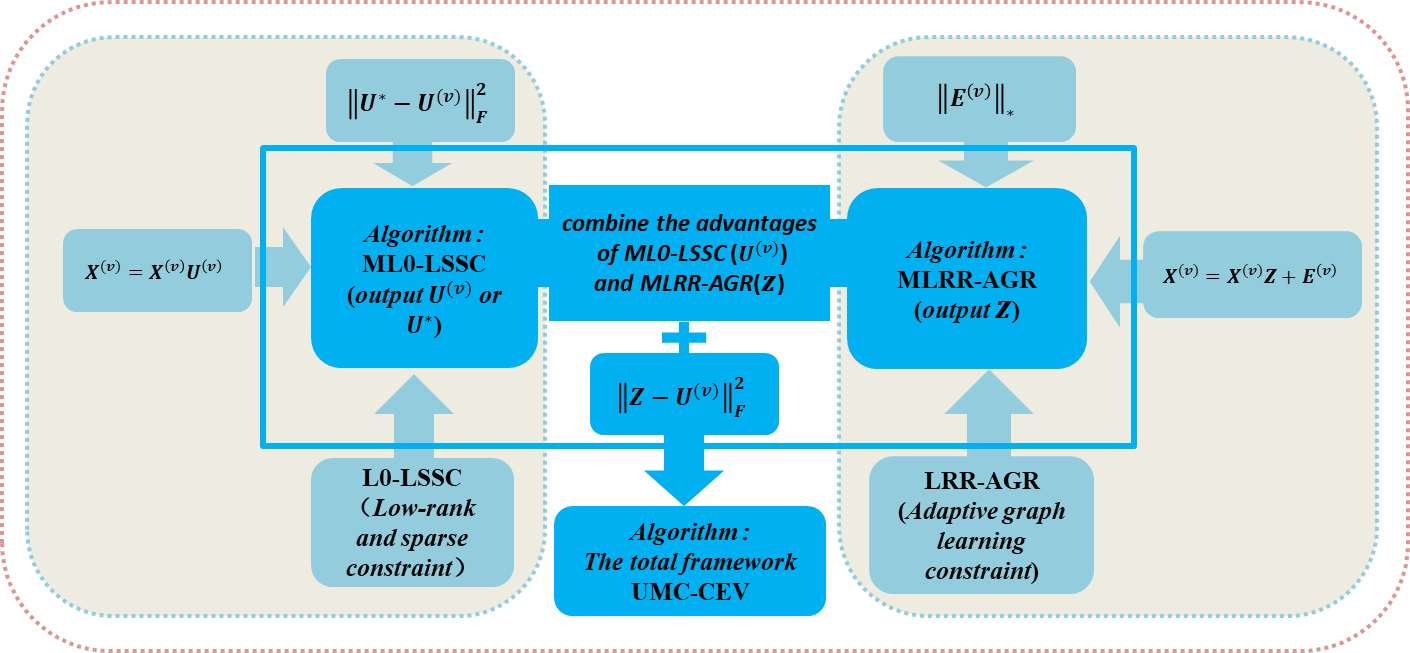}
  \caption{The relationship among ML0-LSSC, MLRR-AGR and UMC-CEV.} 
  \label{fig2:relationship}
\end{figure}

We make full use of the superiority of L0-LSSC and LRR-AGR and then extend them to multi-view clustering. Therefore, this section is divided into three sub-sections, including multi-view generalization and improvement of L0-LSSC, multi-view generalization and improvement of LRR-AGR, and the novel framework (UMC-CEV) based on low-rank SSC and cooperative learning with AGR. The relationship among these three sub-sections is shown in Fig. \ref{fig2:relationship}, and details are given in the following section.

\subsection{Multi-view $L_0$-Motivated Low-rank Sparse Subspace Clustering (ML0-LSSC)}
In an SSC algorithm, we argue that the two constraints of the model (i.e., low rank and sparsity) are critical. These two parts can directly determine the feature extraction of the original data, remove redundant information and reduce the data dimension. L0-LSSC can achieve perfect clustering results when dealing with low-rank and sparse problems. We should make full use of the superiority of the algorithm in low rank and sparseness and generalize them to the multi-view clustering algorithm. Due to the diversity of multi-view data, the diverse information of each view attracts considerable attention. Therefore, we can use the L0-LSSC algorithm to effectively extract the view information by sparse subspace decomposition. In addition, we can effectively integrate various information of each view by cooperative learning. The proposed loss function, which is a modification of L0-LSSC, is as follows:
\begin{equation}
\begin{aligned}
\min _{U^{*}, U^{(v)}} &\sum_{v=1}^{n_{v}} \frac{1}{2}\left\|X^{(v)}-X^{(v)} U^{(v)}\right\|_{F}^{2}+\varphi_{B}\left(\delta\left(U^{(v)}\right)\right)\\
&+\varphi_{B}\left(U^{(v)}\right)+\eta\left\|U^{*}-U^{(v)}\right\|_{F}^{2}, \\
&\text { s.t. } \operatorname{diag}\left(U^{*}\right)=0, \operatorname{diag}\left(U^{(v)}\right)=0
\end{aligned}
\end{equation}

\noindent where the variable $U^\ast$ denotes the global sparse subspace matrix and the parameter $\eta$ trades off the importance of the global sparse subspace matrix. The second and third terms denote low-rank and sparsity constraints, respectively. For simplicity, we use a uniform variable instead of these two terms as follows: 
\begin{equation}
\phi\left(U^{(v)}\right)=\varphi_{{B}}\left(\delta\left(U^{(v)}\right)\right)+\varphi_{B}\left(U^{(v)}\right).
\end{equation}

The fourth term fuses the diversity of each view. Therefore, we can simplify Eq. (4) as:
\begin{equation}
\begin{aligned}
&\min \sum_{v=1}^{n_{v}} \phi\left(U^{(v)}\right)+\eta\left\|U^{*}-U^{(v)}\right\|_{F}^{2}. \\
&\text { s.t. } X^{(v)}=X^{(v)} U^{(v)}, 
\operatorname{diag}\left(U^{*}\right)=0,\operatorname{diag}\left(U^{(v)}\right)=0
\end{aligned}
\end{equation}

\subsection{Multi-view Low-rank Representation with Adaptive Graph Regularization (MLRR-AGR)}
Adaptive graph learning also plays an indispensable role in the field of multi-view clustering. The preservation of the local geometry of each view in sparse subspace decomposition and the retention diversity of each view in fusion are two hot research issues. Hence, adaptive graph learning is able to handle these problems well. In this section, our main objective is to improve low-rank representation with AGR in the multi-view field. First, to remove the redundancy and noise information of the original data, we perform a common sparse subspace decomposition of the original data. The raw data are decomposed into a global sparse subspace and the noise space of each view. Then, we use this decomposed global sparse subspace matrix as the global similarity matrix for AGR and discriminant clustering information. The final optimization model can be given as follows:

\begin{equation}
\begin{aligned}
\min _{Z, E^{(v)}, F^{(v)}} \sum_{v=1}^{n_{v}} \sum_{i, j}^{n}&\left\|x_{i}^{(v)}-x_{j}^{(v)}\right\|_{2}^{2} z_{i j}+\lambda_{1}\left\|E^{(v)}\right\|_{*}\\
&+2 \lambda_{2} \operatorname{tr}\left(\left(F^{(v)}\right)^{T} L_{z} F^{(v)}\right), \\
\end{aligned}
\end{equation}
\begin{align*}
&\text{s.t.} X^{(v)}=X^{(v)} Z+E^{(v)}, \operatorname{diag}(Z)=0,\\
&Z \geq 0, \Sigma_{j} z_{i j}=1,\left(F^{(v)}\right)^{T} F^{(v)}=I
\end{align*}
 
\noindent where $F^{(v)}=\left[{{(f}_1^v)}^T,{{(f}_2^v)}^T,\cdots,{{(f}_n^v)}^T\right]\epsilon R^{n\times c}$, $n$ and $c$ are the numbers of samples and clusters of the original data, respectively, and $tr(\bullet)$ denotes the trace operator. By Eq. (7), the original data matrix can be decomposed into a global sparse subspace matrix and the noise matrix of each view. In this way, this approach is able to effectively integrate the diversity between views. The first term is AGR to preserve the local geometry. The second term is used to constrain the error between individual views and the common view, where the nuclear norm is mainly used to minimize the error matrix. The third term is adaptive clustering discrimination learning between samples. It can measure the inter-class and intra-class discrimination information between samples to a certain extent.

\subsection{The Unified Framework of UMC-CEV}
In the above two sub-sections, we have proposed the improvement of SSC and AGR from single feature space to multi-view feature space. In this section, we effectively combine these two parts to generate an optimal model for comprehensive consideration. First, according to the data matrix decomposition methods of ML0-LSSC and MLRR-AGR, each view has its own unique features and some common features between various views. If the common features are not extracted well, they may become redundant information in the final clustering and depress the final classification performance. Therefore, we need to simultaneously extract the effective common features between views and decompose the optimal diversity of each view. We can learn using Eq. (6) by training the views collaboratively to solve the common information between views and the unique information of each view. Second, we should make full use of the low-rank and sparsity constraints to make our model more robust. Finally, the similarity matrix of the entire model should maintain its AGR characteristics. For a loss function that satisfies the above three conditions, we can define the final model as follows:

\begin{equation}
\begin{aligned}
\min _{Z, E^{(v)}, U^{(v)},F^{(v)}}  \sum_{v=1}^{n_{v}} \sum_{i, j}^{n}\left\|x_{i}^{(v)}-x_{j}^{(v)}\right\|_{2}^{2} z_{i j}+\lambda_{3}\left\|E^{(v)}\right\|_{*}\\
+2 \lambda_{2} \operatorname{tr}\left(\left(F^{(v)}\right)^{T} L_{z} F^{(v)}\right)\\ 
+\sum_{v=1}^{n_{v}} \lambda_{1} \phi\left(U^{(v)}\right)+\eta\left\|Z-U^{(v)}\right\|_{F}^{2},\\
\end{aligned}
\end{equation}

\begin{align*}
\text { s.t. } &X^{(v)}=X^{(v)} Z+E^{(v)}, X^{(v)}=X^{(v)} U^{(v)} \\
&\operatorname{diag}\left(U^{(v)}\right)=0, \operatorname{diag}(Z)=0, Z \geq 0,\\
&\Sigma_{j} z_{i j}=1,\left(F^{(v)}\right)^{T} F^{(v)}=I
\end{align*}

\noindent where parameters $\lambda_1$, $\lambda_2$, $\lambda_3$ and $\eta$ denote penalty factors that trade off the low-rank sparse subspace matrix, the clustering similarity matrix, the error matrix of each view, the discrimination of the global similarity matrix and the similarity of each view, respectively. $L_z$ is the Laplacian matrix, expressed as $L_z=D-(Z+Z^T)/2$, where D is the diagonal matrix, which can be expressed as $D_{ii}=\sum_{j}(z_{ij}+z_{ji})/2$. The first term primarily preserves the local geometric structure of the original data. The second term is the penalty for the decomposition error matrix of each view by using the nuclear norm. Since the nuclear norm is a low-rank constraint on the error matrix, it can be further retained as much as possible by this way. The third term is included to trade off the clustering similarity. We can obtain a very intuitive understanding of the transformation of the following equation:

\begin{equation}
\operatorname{tr}\left(\left(F^{(v)}\right)^{T} L_{z} F^{(v)}\right)=\sum_{i, j}^{n}\left\|f_{i}^{(v)}-f_{j}^{(v)}\right\|_{2}^{2} z_{i j},
\end{equation}

\noindent where $f_i^{(v)}\in R^{c\times1}$ is the clustering index vector, which is an unsupervised learning vector. Eq. (9) uses the similarity matrix as the weight of each sample clustering label to minimize the intra-class error and maximize the inter-class error. The fourth term in Eq. (8) denotes the low-rank and sparsity constraints on the similarity of each view and can effectively extract the complementary information unique to each view. The fifth term in Eq. (8) minimizes the error of the global similarity matrix and view diversity. This allows the global similarity matrix to have more common features for each view. 

\indent To sum up, we propose a unified efficiently unsupervised multi-view clustering framework (i.e., UMC-CEV), that contains two specific multi-view clustering sub-modules (ML0-LSSC and MLRR-AGR). ML0-LSSC algorithm focuses on extracting the unique information of each view and the common view information, while MLRR-AGR algorithm focuses on extracting the cross-view common information between views. To fully combine the advantages of ML0-LSSC and MLRR-AGR, UMC-CEV is proposed by considering frame structure context. This paper focuses on introducing the model structure and optimization process of the algorithm UMC-CEV. UMC-CEV is a new unsupervised multi-view clustering framework with sparse and low-rank structure. A common matrix containing all the common information and multiple pure specific view characteristics with each view characteristic are constructed to squeeze context knowledge. This framework utilizes two different novel sparse matrix decomposition methods to extract the features of the common view and specific views. Furthermore, novel adaptive graph learning, and relevant sparse and low-rank constraints are used to maintain the structural features of the original data.

\section{Optimization \vspace{-0em}}

In this section, we introduce the optimization process of the proposed model. Due to the complexity of the model, to more easily calculate the solutions of variables, we introduce the auxiliary variables $S$, $U_1^{(v)}$ and $U_2^{(v)}$ and then adopt ADMM to solve the optimization problems. The optimization model in Eq. (8) can be converted to:
\begin{equation}
\begin{aligned}
&\min_ {Z, S, E^{(v)}, U^{(v)}, U_{1}^{(v)}, U_{2}^{(v)}, F^{(v)}} \sum_{v=1}^{n_{v}} \sum_{i, j}^{n}\left\|x_{i}^{(v)}-x_{j}^{(v)}\right\|_{2}^{2} z_{i j}\\
&+\sum_{v=1}^{n_{v}} \lambda_{1} \phi\left(U^{(v)}\right)+\lambda_{3}\left\|E^{(v)}\right\|_{*}+2 \lambda_{2} \operatorname{tr}\left(\left(F^{(v)}\right)^{T} L_{z} F^{(v)}\right).\\
\end{aligned}
\vspace{-2ex}
\end{equation}
\begin{align*}
&\text{s.t.} X^{(v)} =X^{(v)} Z+E^{(v)}, X^{(v)}=X^{(v)} U^{(v)}, \\
&Z=S,Z=U^{(v)}=U_{1}^{(v)};\\
&U^{(v)} =U_{2}^{(v)}-\operatorname{diag}\left(U_{2}^{(v)}\right), \operatorname{diag}(S)=0, \\
&S \geq 0, \Sigma_{j} s_{i j}=1,\left(F^{(v)}\right)^{T} F^{(v)}=I
\vspace{-2ex}
\end{align*}

To understand the optimization process of the model more intuitively, we rewrite the model as the following augmented Lagrangian formulation:
\begin{equation}
\begin{aligned}
&L\left(Z, S, E^{(v)}, U^{(v)}, U_{1}^{(v)}, U_{2}^{(v)}\right)=\\&\sum_{v=1}^{n_{v}} \sum_{i, j}^{n}\left\|x_{i}^{(v)}-x_{j}^{(v)}\right\|_{2}^{2} s_{i j}+\sum_{v=1}^{n_{v}} \lambda_{1} \phi\left(U^{(v)}\right)\\&+\lambda_{3}\left\|E^{(v)}\right\|_{*}+2 \lambda_{2} \operatorname{tr}\left(\left(F^{(v)}\right)^{T} L_{s} F^{(v)}\right)\\
&+\left\langle C_{1}^{(v)}, U^{(v)}-U_{1}^{(v)}\right\rangle+\left\langle C_{2}^{(v)}, U^{(v)}-U_{2}^{(v)}+\operatorname{diag}\left(U_{2}^{(v)}\right)\right\rangle \\
\end{aligned}
\nonumber
\end{equation}
\begin{equation}
\begin{aligned}
&+\frac{\mu_{1}}{2}\left\|U^{(v)}-U_{1}^{(v)}\right\|_{F}^{2}+\frac{\mu_{2}}{2}\left\|U^{(v)}-U_{2}^{(v)}+\operatorname{diag}\left(U_{2}^{(v)}\right)\right\|_{F}^{2}\\
&+\frac{\eta}{2}\left(\left\|Z-U^{(v)}\right\|_{F}^{2} +\left\|X^{(v)}-X^{(v)} U^{(v)}\right\|_{F}^{2}\right) \\
&+\frac{\mu}{2}\left(\left\|X^{(v)}-X^{(v)} Z-E^{(v)}\right\|_{F}^{2}+\|Z-S\|_{F}^{2}\right), 
\end{aligned}
\end{equation}

\noindent where $C_1^{(v)}$ and $C_2^{(v)}$ are the coefficient matrices, $\eta$ and $\mu$ are Lagrange multipliers, and $\mu_1$ and $\mu_2$ are positive penalty factors. We can optimize one variable by fixing the other variables in Eq. (11). Therefore, we can obtain the optimal values of all variables $Z$, $S$, $E^{\left(v\right)}$, $U^{\left(v\right)}$, $U_1^{\left(v\right)}$, $U_2^{\left(v\right)}$, and $F^{\left(v\right)}$. The optimization process for these variables is given as follows:

1) Update $E^{(v)}$: When variables $Z$, $S$, $F^{\left(v\right)}$, $U^{\left(v\right)}$,  $U_1^{\left(v\right)}$, and ${\ U}_2^{\left(v\right)}$ are fixed, the objective function in Eq. (11) for $E^{(v)}$ gives the following minimization problem:

\begin{equation}
\underset{E^{(v)}}{\arg \min } \, \lambda_{3}\left\|E^{(v)}\right\|_{*}+\frac{\mu}{2}\left\|X^{(v)}-X^{(v)} Z-E^{(v)}\right\|_{F}^{2}.
\end{equation}

Eq. (12) can be solved by the singular value threshold method. $A=X^{(v)}Z-X^{\left(v\right)}$ can be decomposed by SVD as $U\Sigma V^T$. The optimal solution to Eq. (12) is as follows:
\begin{equation}
E^{(v)}=U S_{\lambda_{3} / \mu}(\Sigma) V^{T}.
\end{equation}

2) Update $Z$: Variables $S$, $E^{\left(v\right)}$, $U^{\left(v\right)}$, $U_1^{\left(v\right)}$,  $U_2^{\left(v\right)}$ and $F^{\left(v\right)}$ should be fixed, and the objective function for $Z$ can be rewritten as follows:

\begin{equation}
\begin{aligned}
L(Z)=\left\|X^{(v)}-X^{(v)} Z-E^{(v)}\right\|_{F}^{2}\\
+\frac{\eta}{2}\left\|Z-U^{(v)}\right\|_{F}^{2}+\frac{\mu}{2}\|Z-S\|_{F}^{2}.
\end{aligned}
\end{equation}

By local derivation of Eq. (14) for ${\partial L}/{\partial Z=0}$, we can obtain the optimal variable $Z$ as follows:

\begin{subequations}
\begin{equation}
Z=L_{3} * \left(\left(X^{(v)}\right)^{T} * L_{1}+L_{2}\right).
\vspace{-2ex}
\end{equation}
\begin{equation}
L_{1}=X^{(v)}-E^{(v)};
\vspace{-2ex}
\end{equation}
\begin{equation}
L_{2}=\eta * U^{(v)}+\mu * S;
\vspace{-2ex}
\end{equation}
\begin{equation}
L_{3}=\left(\left(X^{(v)}\right)^{T} X^{(v)}+(\eta+\mu) I\right)^{-1}.
\end{equation}
\end{subequations}

3) Update $S$: Variables $Z$, $E^{\left(v\right)}$, $U^{\left(v\right)}$, $U_1^{\left(v\right)}$, $U_2^{\left(v\right)}$ and ${\ F}^{\left(v\right)}$ should be fixed, and the objective function for $S$ can be rewritten as follows:

\begin{equation}
\begin{aligned}
\underset{S} {\min } \,  &\sum_{i, j}^{n}\left\|x_{i}^{(v)}-x_{j}^{(v)}\right\|_{2}^{2} s_{i j}+\frac{\mu}{2}\|Z-S\|_{F}^{2}\\
&+2 \lambda_{2} \operatorname{tr}\left(\left(F^{(v)}\right)^{T} L_{s} F^{(v)}\right). \\
\end{aligned}
\end{equation}
\begin{align*}
&\text{s.t.} \operatorname{diag}(\mathrm{S})=0, \mathrm{S} \geq 0, \Sigma_{j} s_{i j}=1
\end{align*}

In Eq. (16), it is difficult to solve for the variable $S$. However, Eq. (16) is equivalent to solving the following minimization problem by adding Eq. (9):

\begin{equation}
\begin{aligned}
\underset{S} {\min } \, &\sum_{i, j}^{n}\left\|x_{i}^{(v)}-x_{j}^{(v)}\right\|_{2}^{2} s_{i j}+\lambda_{2} \sum_{i, j}^{n}\left\|f_{i}^{(v)}-f_{j}^{(v)}\right\|_{2}^{2} s_{i j}\\
&+\frac{\mu}{2}\|Z-S\|_{F}^{2}. \\
&\text{s.t.}\operatorname{diag}(\mathrm{S})=0, \mathrm{S} \geq 0, \Sigma_{j} s_{i j}=1
\end{aligned}
\end{equation}

We further simplify Eq. (17) as:
\begin{equation}
\underset{s_{i}>0, s_{i}I=1, s_{ii}=0} {\min } \, \left\|s_{i}-\left(h_{i}-g_{i}^{(v)} / \mu\right)\right\|_{2}^{2},
\end{equation}

\noindent where $g_{ij}^{(v)}=\lVert x^{(v)}_{i} - x^{(v)}_{j}\rVert_{2}^{2}+\lambda_{2}\lVert f_{i}^{(v)}-f_{j}^{(v)}\rVert_{2}^{2}$, $h_i=z_i$ and $s_i$ are the $i$-th rows of $G^{(v)}$, $H$ and $S$ respectively. Eq. (18) can be effectively solved according to Ref. \cite{[39]zhan2017graph}.

4) Update $F^{(v)}$: Fix $Z$, $S$, $E^{\left(v\right)}$, $U^{\left(v\right)}$, $U_1^{\left(v\right)}$, and $U_2^{\left(v\right)}$; then, the objective function for solving $F^{(v)}$ can be rewritten as follows:

\begin{equation}
\begin{aligned}
\arg \min 2 \lambda_{2} \operatorname{tr}\left(\left(F^{(v)}\right)^{T} L_{s} F^{(v)}\right),\\
\text{s.t.} \left(F^{(v)}\right)^{T} F^{(v)}=I, F^{(v)} \in R^{n \times c}
\end{aligned}
\end{equation}

\noindent where $L_s$ is the Laplacian matrix of the similarity matrix $S$. This is a classic spectral clustering model. In general, we perform eigenvalue decomposition on the model and then take the feature vector corresponding to its top $c$ smallest eigenvalues as $c$ column vectors of the solution $F^{(v)}$. 

5) Update $U^{(v)}$: When variables $Z$, $S$, $F^{\left(v\right)}$, $E^{\left(v\right)}$,  $U_1^{\left(v\right)}$ and $U_2^{\left(v\right)}$ are fixed, the objective function for $U^{(v)}$ is the following minimization problem:

\begin{equation}
\begin{aligned}
\mathrm{L}\left(U^{(v)}\right)&=\min _{U^{(v)}} \frac{1}{2}\left\|X^{(v)}-X^{(v)} U^{(v)}\right\|_{F}^{2}+\frac{\eta}{2}\left\|Z-U^{(v)}\right\|_{F}^{2}\\
&+\frac{\mu_{1}}{2}\left\|U^{(v)}-U_{1}^{(v)}\right\|_{F}^{2}+\left\langle C_{1}^{(v)}, U^{(v)}-U_{1}^{(v)}\right\rangle\\
&+\frac{\mu_{2}}{2}\left\|U^{(v)}-U_{2}^{(v)}+\operatorname{diag}\left(U_{2}^{(v)}\right)\right\|_{F}^{2}\\
&+\left\langle C_{2}^{(v)}, U^{(v)}-U_{2}^{(v)}+\operatorname{diag}\left(U_{2}^{(v)}\right)\right\rangle.
\end{aligned}
\end{equation}

By local derivation of Eq. (20) for $\partial L(U^{\left(v\right)})/\partial \left(U^{\left(v\right)}\right)=0$, the solution of $U^{\left(v\right)}$ can be obtained by:

\begin{subequations}
\begin{equation}
U^{(v)} =B *\left[\left(X^{(v)}\right)^{T} X^{(v)}+M_{1}-M_{2}\right];
\vspace{-1ex}
\end{equation}
\begin{equation}
B =i n v\left(\left(X^{(v)}\right)^{T} X^{(v)}+\left(\mu_{1}+\mu_{2}-\eta\right) I\right).
\vspace{-1ex}
\end{equation}
\end{subequations}
\begin{equation}
M_{1} =\mu_{1} U_{1}^{(v)}+\mu_{2} U_{2}^{(v)}, M_{2}=\eta Z+C_{1}^{(v)}+C_{2}^{(v)}.
\end{equation}

6) Update $U_1^{(v)}$: To solve for this variable, we should fix the other variables $Z$, $S$,  $E^{\left(v\right)}$, ${F^{(v)}}$, $U^{\left(v\right)}$ and $U_2^{\left(v\right)}$. Then, the optimal $U_1^{(v)}$ can be obtained by solving the following minimization problem:

\begin{equation}
\begin{aligned}
\min _{U_{1}^{(v)}} &\lambda_{1} \varphi_{B}\left(\delta\left(U_{1}^{(v)}\right)\right)+\frac{\mu_{1}}{2}\left\|U^{(v)}-U_{1}^{(v)}\right\|_{F}^{2}\\
&+\left\langle C_{1}^{(v)}, U^{(v)}-U_{1}^{(v)}\right\rangle.
\end{aligned}
\end{equation}

Eq. (23) is equivalent to:

\begin{equation}
\begin{aligned}
\min _{U_{1}^{(v)}} &\lambda_{1} \varphi_{B}\left(\delta\left(U_{1}^{(v)}\right)\right)+\\&\frac{\mu_{1}}{2}\left\|U_{1}^{(v)}-\left(\frac{C_{1}^{(v)}}{\mu_{1}}+U^{(v)}\right)\right\|_{F}^{2}.
\end{aligned}
\end{equation}

According to Eq. (19), ${C_1^{\left(v\right)}}/{\mu_1}+U^{\left(v\right)}$ can be decomposed by SVD as $U\Sigma V^T$. Therefore, the closed-form solution of Eq. (24) can be obtained by: 

\begin{equation}
U_{1}^{(v)}=U \theta\left(\Sigma; \frac{\lambda_{1}}{\mu_{1}}, \frac{\lambda_{1}}{\gamma \mu_{1}}\right) V^{T} , 0<\gamma \leq 1,
\end{equation}

\noindent where $\theta(\bullet)$ is the firm threshold function as follows:
\begin{equation}
\theta(x, \lambda, a)=\left\{\begin{array}{cc}
{0} & {\text { if }|x| \leq \lambda} \\
{a(|x|-\lambda) /(a-\lambda) \operatorname{\textit{sign}}(x)} & {\text { if } \lambda <|x| \leq a}. \\
{x} & {\text { if }|x| > a}
\end{array}\right.
\end{equation}

7) Update $U_2^{(v)}$: We should fix the other variables $Z$, $S$, $E^{\left(v\right)}$, ${F^{(v)}}$,  $U^{\left(v\right)}$ and $U_1^{\left(v\right)}$. Then, the optimal $U_2^{(v)}$ can be obtained by solving the following minimization problem:

\begin{equation}
\min _{U_{2}^{(v)}} \lambda_{1} \varphi_{B}\left(U_{2}^{(v)}\right)+\frac{\mu_{2}}{2}\left\|U_{2}^{(v)}-\left(\frac{C_{2}^{(v)}}{\mu_{2}}+U^{(v)}\right)\right\|_{F}^{2},
\end{equation}

\noindent by subtracting the diagonal elements of $U_2^{(v)}$ as follows:
\begin{equation}
U_{2}^{(v)} \leftarrow U_{2}^{(v)}-\operatorname{diag}\left(U_{2}^{(v)}\right).
\end{equation}

Through the above analysis, we can finally obtain the solution of $U_2^{(v)}$ as follows:
\begin{subequations}
\begin{equation}
U_{2}^{(v)}=\theta\left(\frac{C_{2}^{(v)}}{\mu_{2}}+U^{(v)} ; \frac{\lambda_{1}}{\mu_{2}}, \frac{\lambda_{1}}{\gamma \mu_{2}}\right) , 0<\gamma \leq 1;
\end{equation}
\begin{equation}
U_{2}^{(v)} \leftarrow U_{2}^{(v)}-\operatorname{diag}\left(U_{2}^{(v)}\right).
\end{equation}
\end{subequations}

8) Update other variables: Now, we will update the Lagrange multipliers $C_1^{(v)}$, $C_2^{(v)}$, $\mu_1$ and $\mu_2$. These variables can be updated as follows:

\begin{equation}
{C_{1}^{(v)}=C_{1}^{(v)}+\mu_{1}\left(U^{(v)}-U_{1}^{(v)}\right)};
\vspace{-1ex}
\end{equation}
\begin{equation}
{C_{2}^{(v)}=C_{2}^{(v)}+\mu_{2}\left(U^{(v)}-U_{2}^{(v)}-\operatorname{diag}\left(U_{2}^{(v)}\right)\right)}.
\end{equation}
\begin{equation}
\mu_{i}=\min \left(\rho_{1} \mu_{i}, \mu_{i}^{\max }\right), i=1,2.
\end{equation}

\noindent where the parameters $\rho_1$, $\mu_1^{max}$ and $\mu_2^{max}$ are constants.

	When we obtain the global similarity matrix and the individual view similarity matrices through the above optimization method, we use the following formula to combine the two parts: 
\begin{equation}
A=\left(|Z|+\left|Z^{T}\right|\right) / 2+\frac{1}{n_{v}} \sum_{v=1}^{n_{v}}\left(\left|U^{(v)}\right|+\left|\left(U^{(v)}\right)^{T}\right|\right) / 2.
\end{equation}
\indent Then, the optimal global similarity matrix is obtained. After obtaining the optimal similarity matrix, we apply spectral clustering to $A$. The optimization process of the entire model is summarized in Algorithm 1.

\newcommand*{\LongState}[1]{\State
\parbox[t]{\linewidth-\algorithmicindent-\algorithmicindent}{#1\strut}}

 In the overall algorithm optimization process, we firstly analyze the main three optimization sub-modules with relatively high algorithm complexity compared with other variable optimization modules. Such as singular value thresholding (in step 3), the inverse operation (in step 4), and eigen-decomposition (in step 6). These are all classical optimization methods, and we can easily get their algorithm complexity of $O(n^3)$, $O(n^3)$ and $O(cn^2)$, respectively, where $n$ is the number of the samples and $c$ is the number of the clustering. Besides, the computational complexity of $U$ is $O((mn^2+t_1 n^3)$, where $t_1$ is the number of iterations and $m$ is the size of the each sample in step 7. Due to $c \ll m$, we can get the computational complexity of the proposed method as $O(tn_v (n^3+mn^2))$, where $t$ and $n_v$ are the number of iterations and samples views, respectively. Meanwhile, we can easily get the complexity analysis of several comparison methods of MLSSC$(O(tn_v n^3))$, MVGL$(tn_v^2 n^2)$, LRPP-GRR$(O(t(n^3+cn^2)))$, L0-LSSC$(O(tn^3+mn^2 ))$. By comparing the algorithm complexity with these methods, we can see that the proposed algorithm is feasible in theory.

\begin{algorithm}[]
\caption{ADMM for solving the proposed algorithm (UMC-CEV)}
\hspace*{0.02in} {\bf Input:} 
input parameters Dataset $X=\{{X^{(v)}}\}_{v=1}^{n_v}$; parameters $\lambda_1$, $\lambda_2$, and $\lambda_3$; number of clusters $C$; number of views $n_v$\\
\hspace*{0.02in} {\bf Output:} 
output The similarity matrix $A$.\\
\hspace*{0.01in} {\bf Initialization:} 
Using the k-nearest neighbour graph to initialize the similarity matrix $Z$; $S=Z$; using the eigenvalue decomposition of Laplacian matrix $Z$ to initialize matrix $F^{(v)}$. Calculating the initial matrix $U^{(v)}$ by Eq. (21); $U_1^{(v)}=0$, $U_2^{(v)}=0$, $C_1^{(v)}=C_2^{(v)}=0$; $E^{(v)}=0$; $\mu{=0.01,\ \ \mu}_1=1$; $\mu_2=0.1$; $\rho_1=1.2$; $\mu_1^{max}=\mu_2^{max}={10}^6$.
\begin{algorithmic}[1]
\While{not converged } 
	\For{$v=1$ to $n_v$} 
		\State Update variable $E^{(v)}$ by using Eq. (13).
		\State Update variable $Z$ by using Eq. (15a).
		\State Update variable $S$ by solving Eq. (18).
		\State Update variable $F^{(v)}$ by solving Eq. (19).
		\State Update variable $U^{(v)}$ by using Eq. (21a).
		\State Update variable $U_1^{(v)}$ by using Eq. (25).
		\LongState{Update variable $U_2^{(v)}$ by using Eq. (29a) and Eq. (29b).}
		\LongState{Update the other variables $C_1^{\left(v\right)}$,$C_2^{\left(v\right)}$,	               		and	$\mu_1$, $\mu_2$ by using Eq. (30), Eq. (31), and Eq. (32).}	 		 	
	\EndFor
\EndWhile\\
\Return $A$
\end{algorithmic}
\end{algorithm}

\section{Experiments}
This section is divided into several sub-sections: experimental settings, comparison algorithms, evaluation metrics, parameter analysis, clustering results, similarity matrix analysis and convergence analysis. Details are given as follows.

\subsection{Experimental Settings}
To fully consider the superiority of our proposed algorithm, we test our algorithm for four kinds of datasets: face dataset, news article dataset, handwritten digital dataset and textual dataset. The face dataset includes the $ORL\underline{\hspace{0.5em}}mtv$ dataset and the Extended YaleB ($EYaleB$) dataset. The news article dataset is the 3-$sources$ dataset. The handwritten digital dataset is the $uci\underline{\hspace{0.5em}}digit$ dataset, and the textual dataset is $BBCSport$. The statistics of the five real-world datasets are summarized in Table \ref{tb:Statistics}.

\paragraph{ORL\underline{\hspace{0.5em}}mtv dataset}
There are 10 different grey-scale images for 40 different themes. Images are taken under different conditions, such as different lighting, different facial expressions, and different facial details. In this experiment, we used three views to evaluate the algorithm.
\paragraph{Extended YaleB dataset}
There are 38 individuals and approximately 64 near-frontal images in this face dataset. These images are under different illuminations. In general, other experiments have used the first 10 classes and 64 near-frontal images, for a total of 640 samples. However, we used 38 individuals and 29 near-frontal images in our experiment, for a total of 1102 samples.
\paragraph{3-sources  dataset}
This is a collection of news stories collected from three online news sources (BBC, Reuters and The Guardian). All articles are indicated by the "word bag". For 948 articles, we used 169 topic classes with a placeholder for each article in all three datasets.
\paragraph{uci\underline{\hspace{0.5em}}digit  dataset}
This dataset contains 2000 handwritten digit (0-9) examples extracted from the Dutch utility map. There are 200 examples in each class, and each example has six feature sets. Following the experiment in \cite{[44]brbic2018multi}, we used three feature sets: Fourier coefficients of 76 character shapes, 216 profile correlations, and 64 Karhunen-Love coefficients.
\paragraph{BBCSport dataset}
This dataset contains the latest articles in the five subject areas of the 2004-2005 BBC Sport website. This textual dataset has 544 documents. In our experiment, we used 116 samples and 5 classes. 
\begin{table}[h]
	\centering
	\caption{Statistics of the three datasets}
	\vspace{1ex}
	\begin{tabular}{c c c c}
		\hline
		Dataset & Sample & Views & Clusters \\ 
		\hline
		\centering ORL\underline{\hspace{0.5em}}mtv & 400 & 3 & 40\\
		\centering EYaleB & 1102 & 2 & 38\\
		\centering 3-sources & 169 & 3 & 6\\
		\centering uci\underline{\hspace{0.5em}}digit & 2000 & 3 & 10\\
		\centering BBCSport & 116 & 3 & 5\\
		\hline
	\end{tabular} 
	
	\label{tb:Statistics}
\end{table}

\subsection{Comparison Algorithms}
In this sub-section, we compare our proposed algorithm with several related state-of-the-art algorithms, including DiMSC \cite{[24]cao2015diversity}, latent multi-view subspace clustering (LMSC) \cite{[45]zhang2017latent}, exclusivity-consistency regularized multi-view subspace clustering (ECRMSC) \cite{[25]wang2017exclusivity}, multi-view low-rank sparse subspace clustering (MLSSC) \cite{[44]brbic2018multi}, graph learning for multi-view clustering (MVGL) \cite{[29]zhan2018multiview}, low-rank representation with adaptive graph regularization (LRPP-AGR) \cite{[40]wen2018low}, and L0-LSSC \cite{[36]brbic2018}. Detailed descriptions of these state-of-the-art algorithms are given as follows:
\paragraph{DiMSC}
This approach mostly concentrates on enhancing the performance of multi-view clustering by exploring additional information between multi-view features. The HSIC is used to explore the diversity information between multi-view features.
\paragraph{LMSC}
This method consists of finding a potential sparse representation subspace and then reconstructing the data simultaneously based on the learned potential subspace. The method uses the complementarity of multiple view feature spaces to obtain the best view features. This algorithm imposes low-rank constraints on the noise matrix and the sparse subspace matrix.
\paragraph{ECRMSC}
This algorithm is mainly divided into two steps: subspace learning and spectral clustering. It uses the $L_1$-norm as a constraint to extract the complementary information between different sparse subspace matrices. 
\paragraph{MLSSC}
Multi-view low-rank sparse subspace clustering is a classic low-rank sparse subspace method that uses the nuclear norm and $L_1$-norm to constrain the low rank and sparsity of the sparse subspace matrix. In addition, this method obtains complementary information between views by means of the mutual derivation of views.
\paragraph{MVGL}
 This is an adaptive graph learning method. By optimizing the similarity graph matrix, the optimal similarity factor matrix of each view is obtained. Finally, these similarity factor matrices are weighted and fused by the idea of cooperative representation, thus obtaining the most excellent global similarity graph matrix.
\paragraph{LRPP-GRR}
This is a novel method based on AGR. It introduces the original data matrix local geometric structure and the cluster label matrix to discriminate between these two manifold learning constraints. In addition, this method uses the nuclear norm and $L_1$-norm to constrain the sparse subspace matrix and the noisy matrix, respectively.
\paragraph{L0-LSSC}
This is a novel method of low-rank sparse subspace clustering. Considering the deficiencies of the nuclear norm and $L_1$-norm, this method uses a new low-rank and sparse method.

\subsection{Evaluation Metrics}
There are seven evaluation metrics in our paper: clustering accuracy(ACC), normalized information(NMI), purity, precision(P), recall(R), F-score(F) and adjusted rand index(AR).

\begin{figure}[]
\centering

\subfigure[]{
\begin{minipage}[t]{0.5\linewidth}
\centering
\includegraphics[width=1.9in]{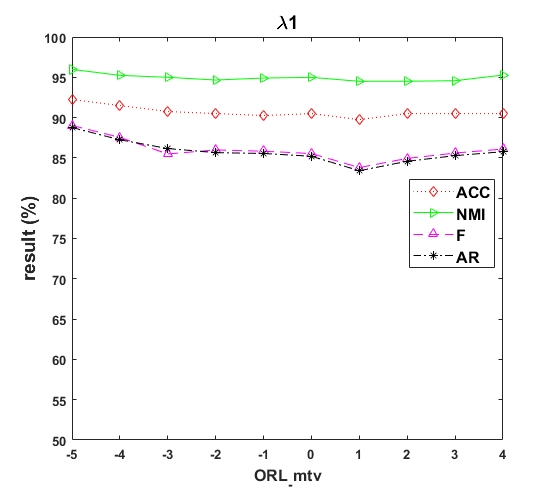}
\end{minipage}%
}%
\subfigure[]{
\begin{minipage}[t]{0.5\linewidth}
\centering
\includegraphics[width=1.9in]{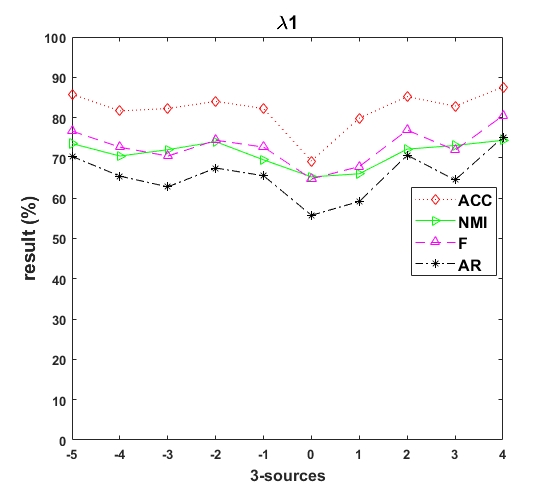}
\end{minipage}%
}%
                               
\subfigure[]{
\begin{minipage}[t]{0.5\linewidth}
\centering
\includegraphics[width=1.9in]{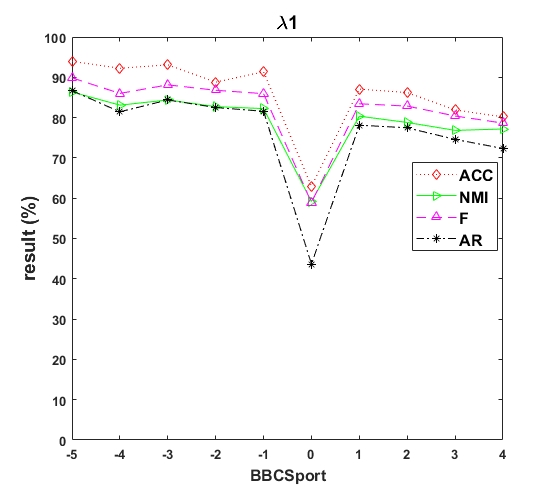}
\end{minipage}%
}%
\subfigure[]{
\begin{minipage}[t]{0.5\linewidth}
\centering
\includegraphics[width=1.9in]{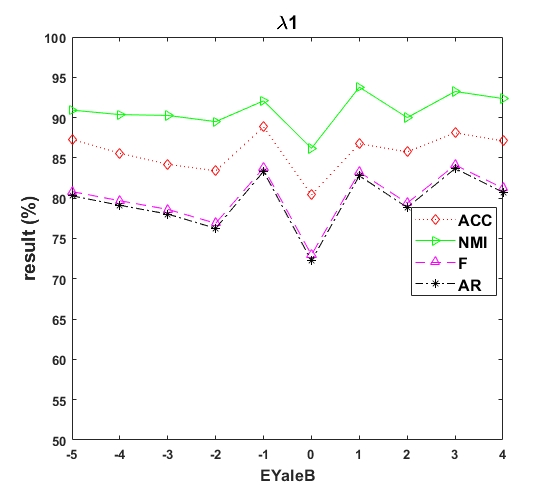}
\end{minipage}%
}%

\subfigure[]{
\begin{minipage}[t]{0.5\linewidth}
\centering
\includegraphics[width=1.9in]{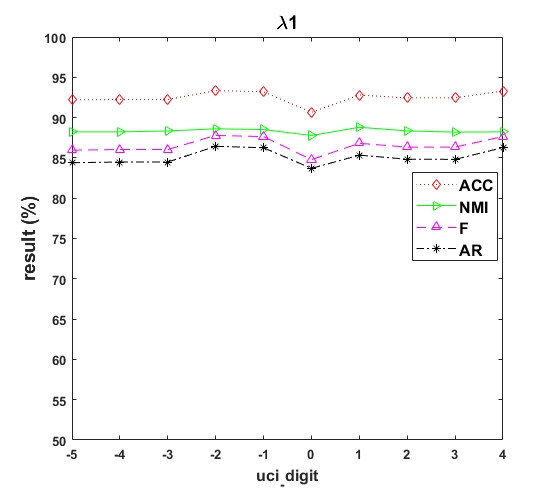}
\end{minipage}%
}%

\centering
\caption{The results according to different values of $\lambda_1$ when $\lambda_2$ and $\lambda_3$ are fixed. Four clustering evaluation indicators (ACC, NMI, F-score and AR) are selected for better visual effect, and we performed logarithmic processing on the parameters ($ln({\lambda_1}/{2})$). }
\label{fig3:results1}
\end{figure}

\begin{figure}
\centering

\subfigure[]{
\begin{minipage}[t]{0.5\linewidth}
\centering
\includegraphics[width=1.9in]{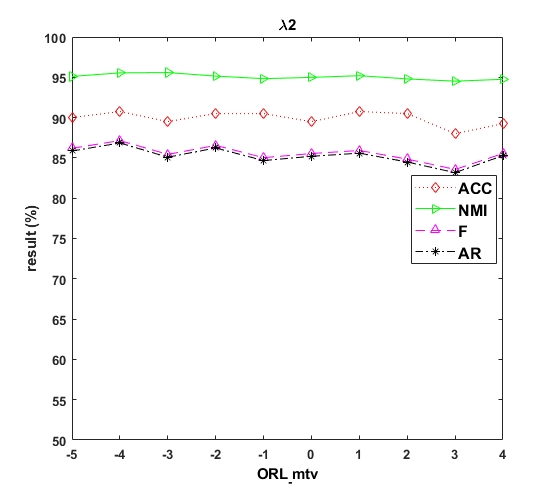}
\end{minipage}%
}%
\subfigure[]{
\begin{minipage}[t]{0.5\linewidth}
\centering
\includegraphics[width=1.9in]{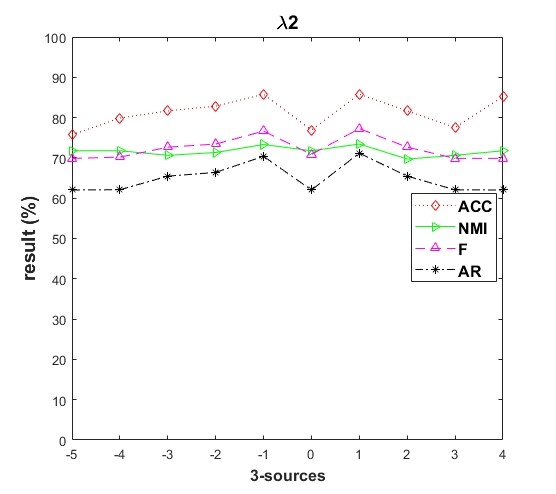}
\end{minipage}%
}%
                               
\subfigure[]{
\begin{minipage}[t]{0.5\linewidth}
\centering
\includegraphics[width=1.9in]{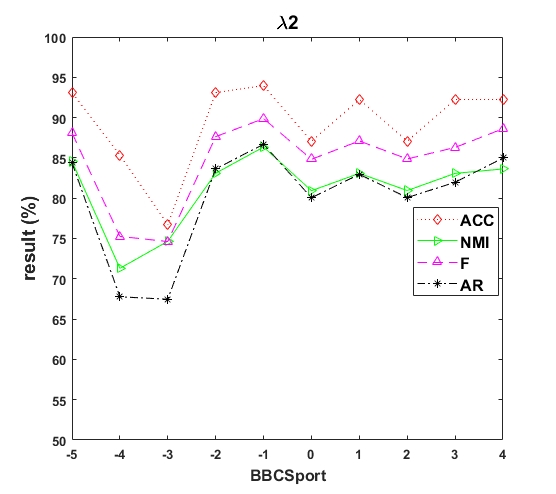}
\end{minipage}%
}%
\subfigure[]{
\begin{minipage}[t]{0.5\linewidth}
\centering
\includegraphics[width=1.9in]{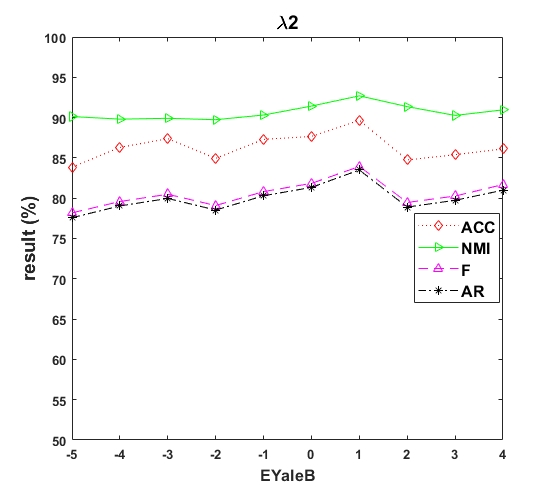}
\end{minipage}%
}%

\subfigure[]{
\begin{minipage}[t]{0.5\linewidth}
\centering
\includegraphics[width=1.9in]{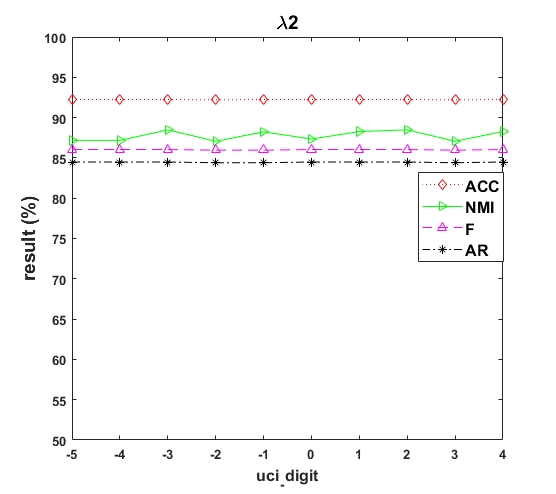}
\end{minipage}%
}%

\centering
\caption{The results according to different values of $\lambda_2$ when $\lambda_1$ and $\lambda_3$ are fixed. Four clustering evaluation indicators (ACC, NMI, F-score and AR) are selected for better visual effect, and we performed logarithmic processing on the parameters ($ln({\lambda_2}/{2})$). }
\label{fig4:results2}
\end{figure}

\begin{figure}[]
\centering

\subfigure[]{
\begin{minipage}[t]{0.5\linewidth}
\centering
\includegraphics[width=1.9in]{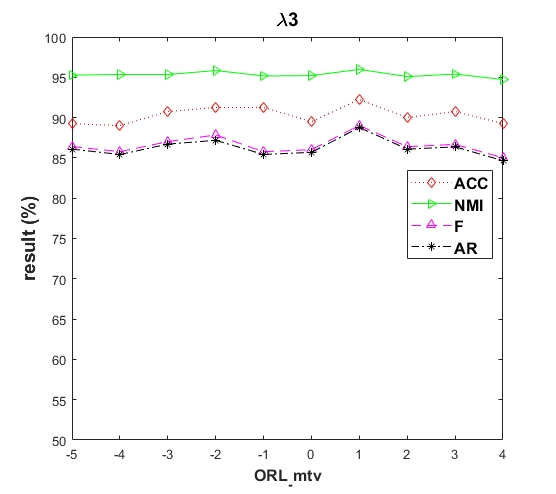}
\end{minipage}%
}%
\subfigure[]{
\begin{minipage}[t]{0.5\linewidth}
\centering
\includegraphics[width=1.9in]{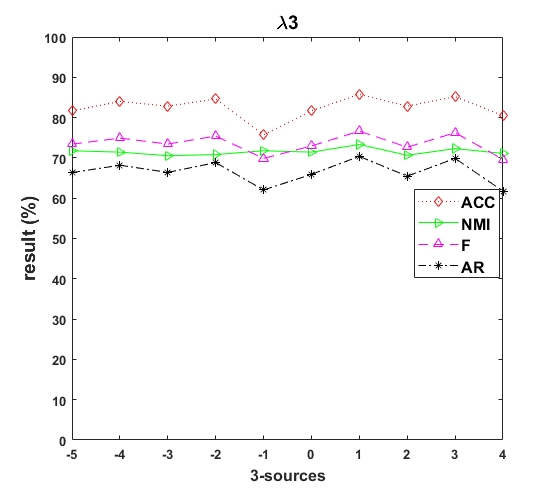}
\end{minipage}%
}%
                               
\subfigure[]{
\begin{minipage}[t]{0.5\linewidth}
\centering
\includegraphics[width=1.9in]{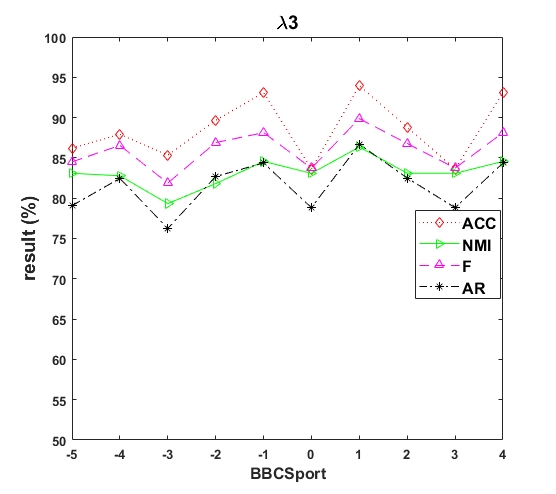}
\end{minipage}%
}%
\subfigure[]{
\begin{minipage}[t]{0.5\linewidth}
\centering
\includegraphics[width=1.9in]{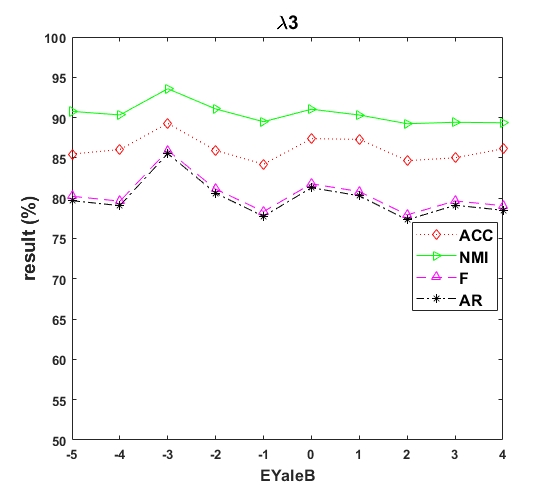}
\end{minipage}%
}%

\subfigure[]{
\begin{minipage}[t]{0.5\linewidth}
\centering
\includegraphics[width=1.9in]{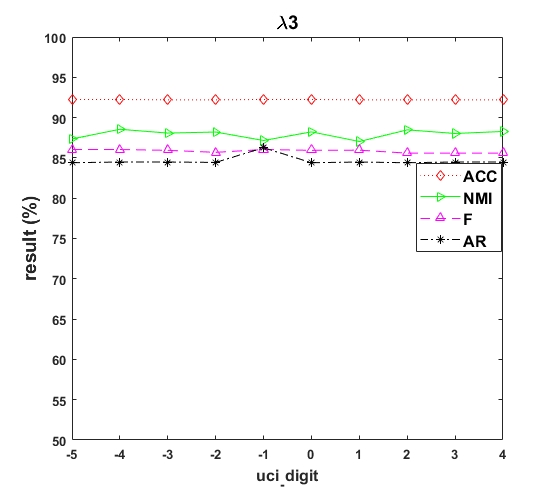}
\end{minipage}%
}%

\caption{The results according to different values of $\lambda_3$ when $\lambda_1$ and $\lambda_2$ are fixed. Four clustering evaluation indicators (ACC, NMI, F-score and AR) are selected for better visual effect, and we performed logarithmic processing on the parameters ($ln({\lambda_3}/{2})$). }
\label{fig5:results3}
\end{figure}

\subsection{Parameter Analysis}
According to Algorithm 1, there are three parameters in our algorithm that need to be determined, i.e., the trade off parameters $\lambda_1$,${\ \lambda}_2$ and ${\ \lambda}_3$. These trade off parameters denote the low-rank constraint on the noisy matrix of each view, the discrimination constraint on the global matrix cluster, and the low-rank and sparsity constraints on each view, respectively. In this sub-section, we analyse the sensitivity of these three sensitive parameters in different datasets. In our experiments, the three parameters are defined within a uniform scope. We choose ten digits for the three parameter values in the range of $[0.00002,0.0002, \cdots, 2, \cdots, 200,2000]$. Figs.\ref{fig3:results1},\ref{fig4:results2},\ref{fig5:results3} show the clustering results according to different values of these three parameters $\lambda_1$,${\ \lambda}_2$ and ${\ \lambda}_3$, respectively. The results for the ORL\underline{\hspace{0.5em}}mtv and uci\underline{\hspace{0.5em}}digit datasets are relatively stable for different values of $\lambda_1$,${\ \lambda}_2$ and ${\ \lambda}_3$. The maximum and minimum accuracies are 92.75\% and 88.75\% 
for the ORL\underline{\hspace{0.5em}}mtv dataset, respectively, and the error between the maximum and minimum accuracies for the uci\underline{\hspace{0.5em}}digit dataset is very small, i.e., 1.15\%. Especially for parameter $\lambda_2$ in Fig. \ref{fig4:results2}, the curves of ACC, NMI and AR in the uci\underline{\hspace{0.5em}}digit dataset are basically unchanged. This is because the uci\underline{\hspace{0.5em}}digit dataset has a large number of samples for each class, and the pictures are handwritten numbers that are easy to distinguish. Therefore, the role of discriminative constraints between different classes is not very large. However, the 3-sources, BBCSport and EYaleB datasets are sensitive to the parameters $\lambda_1$,${\ \lambda}_2$ and ${\ \lambda}_3$. When parameter $\lambda_1=2$, the experimental results for the datasets (3-sources, BBCSport and EYaleB) exhibit particularly obvious fluctuations and are very poor. Thus, it is very important to apply low-rank and sparse constraints in each view. Although the best results can be obtained by constantly adjusting the values of the parameters for different datasets, to guarantee the generality of the proposed algorithm, we use uniform settings for the three parameters in all datasets. We set these three parameters as $\lambda_1=2e-5$, $\lambda_2=2e-1$ and $\lambda_3=2$ for all datasets. Using these values for the parameters, the experimental results are still very good in all datasets.

\begin{table*}[]
	\centering
	\caption{Experimental results (mean $\pm$ standard deviation) 
			ORL\underline{\hspace{0.5em}}mtv for the dataset.}
	\begin{tabular}{c c c c c c c c}
		\hline
		Methods & ACC & NMI & Purity & P & R & F & AR \\ 
		\hline
		LRPP-GRR\underline{\hspace{0.5em}}bestview & $59.50\pm 0.01$ & $75.00\pm 0.06$ & 			$67.50\pm 0.02$ & $42.34\pm 0.07$ & $50.67\pm 0.02$ & $46.13\pm 0.01$ & $43.34\pm 			0.03$\\
		L0-LSSC\underline{\hspace{0.5em}}bestview & $82.24\pm 1.25$ & $93.53\pm 0.66$ & 			$88.94\pm 1.36$ & $69.74\pm 2.16$ & $85.64\pm 1.59$ & $76.85\pm 1.35$ & $76.26\pm 			1.42$\\
		Ours\underline{\hspace{0.5em}}bestview & $88.63\pm 0.73$ & $94.73\pm 0.39$ & 				$91.63\pm 0.68$ & $82.21\pm 1.27$ & $87.62\pm 0.57$ & $84.75\pm 0.98$ & $84.39\pm 			0.93$\\
		\hline
		MLSSC & $70.75\pm 3.55$ & $84.46\pm 1.83$ & $74.50\pm 2.99$ & $58.19\pm 3.97$ & 			$64.33\pm 3.60$ & $61.11\pm 3.76$ & $60.16\pm 3.87$\\
		MVGL & $76.50\pm 0.22$ & $89.89\pm 1.20$ & $86.25\pm 0.34$ & $41.87\pm 0.13$ & 				$81.39\pm 0.34$ & $55.29\pm 1.22$ & $53.92\pm 1.30$\\
		LMSC & $81.75\pm 2.54$ & $93.10\pm 1.15$ & $87.25\pm 1.93$ & $70.32\pm 1.74$ & 				$82.06\pm 1.70$ & $75.83\pm 2.10$ & $72.99\pm 2.19$\\
		DiMSC & $83.25\pm 1.20$ & $93.11\pm 1.40$ & $89.25\pm 2.92$ & $73.96\pm 3.72$ & 			$84.11\pm 3.67$ & $78.71\pm 3.37$ & $78.19\pm 3.45$\\
		ECRMSC & $85.41\pm 1.10$ & $94.70\pm 0.90$ & $91.75\pm 1.21$ & $78.30\pm 0.82$ & 			$85.95\pm 1.21$ & $82.10\pm 1.50$ & $81.01\pm 1.12$\\
		MLRPP-GRR & $83.80\pm 2.48$ & $92.90\pm 0.88$ & $88.82\pm 1.62$ & $75.47\pm 3.24$ 			& $83.61\pm 2.00$ & $79.31\pm 2.40$ & $78.80\pm 2.47$\\
		ML0-LSSC & $85.00\pm 2.63$ & $94.07\pm 0.88$ & $90.50\pm 1.70$ & $75.28\pm 4.43$ 			& $87.11\pm 1.93$ & $81.48\pm 3.24$ & $79.99\pm 3.34$\\
		Ours & \bm{$91.25\pm 0.85$} & \bm{$95.56\pm 0.28$} & \bm{$93.30\pm 0.54$} & 				\bm{$85.40\pm 1.33$} & \bm{$89.86\pm 0.75$} & \bm{$87.30\pm 0.92$} & 						\bm{$87.00\pm 0.95$}\\	
		\hline
	\end{tabular} 
	\label{tb2:ORL}
\vspace{-1.5em}
\end{table*}

\begin{table*}[]
	\centering
	\caption{Experimental results (mean $\pm$ standard deviation) 
			 for the 3-sources dataset.}
	\begin{tabular}{c c c c c c c c}
		\hline
		Methods & ACC & NMI & Purity & P & R & F & AR \\ 
		\hline
		LRPP-GRR\underline{\hspace{0.5em}}bestview & $43.40\pm0.02$ & $36.52\pm0.00$ & 				$75.47\pm0.03$ & $33.52\pm0.01$ & $56.23\pm0.05$ & $42.00\pm0.07$ & $37.40\pm0.07$\\
		L0-LSSC\underline{\hspace{0.5em}}bestview & $76.92\pm1.25$ & $68.45\pm2.21$ & 				$81.66\pm1.96$ & $75.81\pm1.56$ & $69.69\pm1.45$ & $71.71\pm1.89$ & $63.47\pm2.21$\\
		Ours\underline{\hspace{0.5em}}bestview & $83.31\pm1.46$ & $75.25\pm2.25$ & 					$84.91\pm2.14$ & $83.49\pm1.89$ & $70.25\pm1.86$ & $75.89\pm2.25$ & $69.33\pm2.15$\\
		\hline
		MLSSC & $75.15\pm2.07$ & $63.06\pm1.34$ & $79.88\pm2.93$ & $76.29\pm2.06$ & 				$68.48\pm2.19$ & $72.09\pm2.47$ & $64.22\pm2.93$\\
		MVGL & $77.51\pm0.22$ & $67.21\pm0.56$ & $85.57\pm0.00$ & $68.75\pm0.13$ & 					$63.24\pm0.22$ & $65.76\pm0.56$ & $57.23\pm0.71$\\
		LMSC & $76.92\pm2.62$ & $69.59\pm1.17$ & $83.43\pm2.23$ & $76.69\pm1.37$ & 					$67.32\pm2.56$ & $71.70\pm1.93$ & $63.85\pm2.70$\\
		DiMSC & $81.66\pm2.11$ & $69.94\pm1.64$ & $81.66\pm2.45$ & $82.33\pm2.22$ & 				$70.11\pm2.51$ & $75.69\pm2.97$ & $69.07\pm2.14$\\
		ECRMSC & $80.47\pm0.00$ & $70.27\pm0.00$ & $70.47\pm0.00$ & $76.50\pm0.00$ & 				$62.75\pm0.00$ & $68.94\pm0.00$ & $60.71\pm0.00$\\
		MLRPP-GRR & $82.25\pm2.18$ & $72.70\pm2.38$ & $83.21\pm1.89$ & $82.10\pm1.68$ & 			$71.05\pm2.09$ & $75.18\pm2.15$ & $68.22\pm2.37$\\
		ML0-LSSC & $82.25\pm2.74$ & $72.05\pm1.42$ & $82.84\pm2.70$ & $81.68\pm2.36$ & 				$70.81\pm1.59$ & $72.99\pm2.73$ & $65.63\pm2.43$\\
		Ours & \bm{$85.98\pm1.71$} & \bm{$75.64\pm1.12$} & \bm{$86.27\pm1.61$} & 					\bm{$85.79\pm1.49$} & \bm{$72.29\pm2.89$} & \bm{$78.42\pm2.37$} & 							\bm{$72.61\pm2.92$}\\
		\hline
	\end{tabular} 
	\label{tb3:3-sources}
\vspace{-1.5em}
\end{table*}

\begin{table*}[]
	\centering
	\caption{Experimental results (mean $\pm$ standard deviation) 
			  for the BBCSport dataset.}
	\begin{tabular}{c c c c c c c c}
		\hline
		Methods & ACC & NMI & Purity & P & R & F & AR \\ 
		\hline
		LRPP-GRR\underline{\hspace{0.5em}}bestview & $68.97\pm0.02$ & $53.72\pm0.00$ & 					$73.28\pm0.02$ & $51.28\pm0.00$ & $58.48\pm0.01$ & $54.64\pm0.01$ & $38.85\pm0.00$\\
		L0-LSSC\underline{\hspace{0.5em}}bestview & $81.90\pm0.01$ & $68.46\pm0.00$ & 					$81.90\pm0.03$ & $78.07\pm0.00$ & $73.64\pm0.07$ & $72.97\pm0.11$ & $65.06\pm0.20$\\
		Ours\underline{\hspace{0.5em}}bestview & $82.16\pm0.41$ & $82.69\pm0.00$ & 						$86.47\pm0.41$ & $87.64\pm0.80$ & $79.49\pm0.29$ & $82.85\pm0.20$ & $77.68\pm0.07$\\
		\hline
		MLSSC & $63.79\pm2.38$ & $47.48\pm2.55$ & $81.03\pm2.33$ & $52.53\pm2.88$ & 					$66.34\pm2.47$ & $54.89\pm2.17$ & $39.59\pm2.58$\\
		MVGL & $76..28\pm0.17$ & $72.76\pm0.11$ & $81.28\pm0.22$ & $70.44\pm0.11$ & 					$74.62\pm0.13$ & $72.16\pm0.17$ & $63.74\pm0.28$\\
		LMSC & $82.76\pm2.83$ & $78.02\pm1.92$ & $82.21\pm1.46$ & $82.61\pm2.08$ & 						$77.60\pm1.97$ & $80.03\pm2.63$ & $73.90\pm2.30$\\
		DiMSC & $85.34\pm2.70$ & $73.34\pm1.80$ & $85.34\pm2.16$ & $79.75\pm2.45$ & 					$74.13\pm2.96$ & $74.32\pm1.35$ & $66.13\pm2.16$\\
		ECRMSC & $81.03\pm0.00$ & $73.31\pm0.00$ & $81.03\pm0.00$ & $79.36\pm0.00$ & 					$67.82\pm0.00$ & $73.14\pm0.00$ & $65.42\pm0.00$\\
		MLRPP-GRR & $85.76\pm2.23$ & $82.66\pm2.50$ & $85.34\pm1.93$ & $86.85\pm2.14$ & 				$80.88\pm3.36$ & $83.12\pm2.42$ & $77.91\pm2.42$\\
		ML0-LSSC & $84.79\pm2.25$ & $77.41\pm1.99$ & $88.79\pm1.43$ & $83.76\pm2.02$ & 					$81.00\pm2.72$ & $82.14\pm1.60$ & $72.98\pm2.00$\\
		Ours & \bm{$91.90\pm1.70$} & \bm{$84.62\pm1.39$} & \bm{$92.93\pm0.62$} & 						\bm{$89.28\pm1.56$} & \bm{$87.42\pm0.62$} & \bm{$88.20\pm1.28$} & 							\bm{$84.47\pm1.73$}\\
		\hline
	\end{tabular} 
	\label{tb4:BBCSport}
\vspace{-1.5em}
\end{table*}

\begin{table*}[]
	\centering
	\caption{Experimental results (mean $\pm$ standard deviation) 
			  for the EYaleB dataset.}
	\begin{tabular}{c c c c c c c c}
		\hline
		Methods & ACC & NMI & Purity & P & R & F & AR \\ 
		\hline
		LRPP-GRR\underline{\hspace{0.5em}}bestview & $61.38\pm0.00$ & $68.00\pm0.00$ & 					$74.48\pm0.00$ & $58.53\pm0.00$ & $66.11\pm0.00$ & $62.09\pm0.00$ & $52.23\pm0.00$\\
		L0-LSSC\underline{\hspace{0.5em}}bestview & $74.68\pm1.18$ & $81.02\pm0.60$ & 					$81.76\pm0.28$ & $56.56\pm1.25$ & $67.31\pm0.30$ & $61.47\pm1.30$ & $60.37\pm0.89$\\
		Ours\underline{\hspace{0.5em}}bestview & $87.01\pm1.70$ & $91.78\pm0.49$ & 						$92.40\pm0.46$ & $77.84\pm1.96$ & $85.74\pm0.65$ & $81.29\pm1.58$ & $80.77\pm1.66$\\
		\hline
		MLSSC & $66.06\pm3.68$ & $73.73\pm2.33$ & $71.23\pm3.01$ & $48.15\pm3.60$ & 					$55.89\pm3.36$ & $51.73\pm3.45$ & $50.37\pm3.56$\\
		MVGL & $57.12\pm0.11$ & $56.03\pm0.00$ & $61.25\pm0.27$ & $48.07\pm0.00$ & 						$47.75\pm0.00$ & $47.49\pm0.00$ & $45.95\pm0.35$\\
		LMSC & $73.61\pm3.51$ & $83.98\pm1.75$ & $85.06\pm1.71$ & $65.30\pm3.98$ & 						$73.78\pm2.55$ & $69.26\pm3.24$ & $68.40\pm3.34$\\
		DiMSC & $62.32\pm1.57$ & $64.44\pm1.90$ & $63.54\pm1.96$ & $51.64\pm1.24$ & 					$56.52\pm1.35$ & $53.57\pm1.29$ & $47.85\pm1.33$\\
		ECRMSC & $78.98\pm0.00$ & $75.95\pm0.00$ & $74.56\pm0.00$ & $51.32\pm0.00$ & 					$71.85\pm0.00$ & $59.78\pm0.00$ & $54.44\pm0.00$\\
		MLRPP-GRR & $73.91\pm2.70$ & $81.97\pm1.64$ & $80.65\pm1.94$ & $63.54\pm3.03$ & 				$69.41\pm2.44$ & $66.34\pm2.72$ & $65.42\pm2.80$\\
		ML0-LSSC & $73.62\pm4.00$ & $82.51\pm1.96$ & $83.26\pm1.99$ & $63.33\pm4.31$ & 					$71.30\pm2.98$ & $67.05\pm3.61$ & $66.14\pm3.73$\\
		Ours & \bm{$89.04\pm1.88$} & \bm{$93.48\pm1.27$} & \bm{$92.93\pm1.37$} & 					\bm{$82.67\pm2.91$} & \bm{$87.86\pm2.36$} & \bm{$84.84\pm2.58$} & 							\bm{$84.43\pm2.65$}\\
		\hline
	\end{tabular} 
	\label{tb5:EYaleB}
\vspace{-1.5em}
\end{table*}

\begin{table*}[]
	\centering
	\caption{Experimental results (mean $\pm$ standard deviation) 
			  for the uci\underline{\hspace{0.5em}}digit  dataset.}
	\begin{tabular}{c c c c c c c c}
		\hline
		Methods & ACC & NMI & Purity & P & R & F & AR \\ 
		\hline
		LRPP-GRR\underline{\hspace{0.5em}}bestview  & $62.25\pm0.02 $ & $70.97\pm0.00 $ & 				$76.90\pm0.00 $ & $57.13\pm0.01 $ & $67.14\pm0.02 $ & $61.73\pm0.01 $ & $57.12\pm0.01$\\
		L0-LSSC\underline{\hspace{0.5em}}bestview  & $83.20\pm2.32 $ & $75.98\pm2.57 $ & 				$83.20\pm2.58 $ & $70.32\pm1.85 $ & $72.77\pm2.46 $ & $71.52\pm2.46 $ & $68.32\pm2.77$\\
		Ours\underline{\hspace{0.5em}}bestview & $85.80\pm0.00 $ & $80.95\pm1.27 $ & 					$86.91\pm1.03 $ & $74.09\pm2.23 $ & $78.45\pm1.47 $ & $75.92\pm2.21 $ & $73.17\pm2.47$\\
		\hline
		MLSSC  & $87.63\pm2.00 $ & $85.42\pm1.67 $ & $91.62\pm1.54 $ & $81.86\pm2.23 $ & 				$86.42\pm1.67 $ & $83.11\pm2.05 $ & $82.16\pm2.58$\\
		MVGL  & $85.30\pm0.00 $ & $88.21\pm0.30 $ & $94.00\pm0.03 $ & $77.17\pm0.27 $ & 				$89.09\pm0.60 $ & $82.96\pm0.22 $ & $82.26\pm0.45$\\
		LMSC  & $85.70\pm2.70 $ & $77.45\pm1.23 $ & $85.70\pm1.31 $ & $74.58\pm1.68 $ & 				$75.73\pm1.29 $ & $75.15\pm1.32 $ & $72.38\pm1.71$\\
		DiMSC  & $54.00\pm2.70 $ & $40.48\pm1.42 $ & $54.10\pm1.90 $ & $37.17\pm1.57 $ & 				$37.73\pm1.55 $ & $37.39\pm1.55 $ & $30.42\pm1.73$\\
		ECRMSC & $90.35\pm0.00 $ & $84.45\pm0.00 $ & $90.35\pm0.00 $ & $82.64\pm0.00 $ & 				$83.33\pm0.00 $ & $82.98\pm0.00 $ & $81.09\pm0.00$\\
		MLRPP-GRR  & $85.71\pm2.98 $ & $84.16\pm0.89 $ & $89.04\pm1.08 $ & $78.63\pm2.99 $ & 			$82.28\pm1.31 $ & $80.39\pm1.78 $ & $78.16\pm2.03$\\
		ML0-LSSC & $82.07\pm3.39 $ & $82.76\pm1.46 $ & $89.30\pm1.10 $ & $76.88\pm2.55 $ & 				$82.22\pm1.57 $ & $79.45\pm2.01 $ & $77.36\pm1.57$\\
		Ours & \bm{$92.25\pm1.20 $} & \bm{$88.34\pm1.38 $} & \bm{$94.45\pm1.04 $} & 				\bm{$85.69\pm2.31 $} & \bm{$89.56\pm1.48 $} & \bm{$86.04\pm1.23 $} & 						\bm{$84.49\pm2.30$}\\
		\hline
	\end{tabular} 
	\label{tb6:uci}
\end{table*}

\begin{figure}[]
\vspace{-0.8cm}  
\setlength{\abovecaptionskip}{-0.cm}  
\setlength{\belowcaptionskip}{-0.cm}   
\centering

\subfigure[]{
\begin{minipage}[t]{0.5\linewidth}
\centering
\includegraphics[width=1.9in]{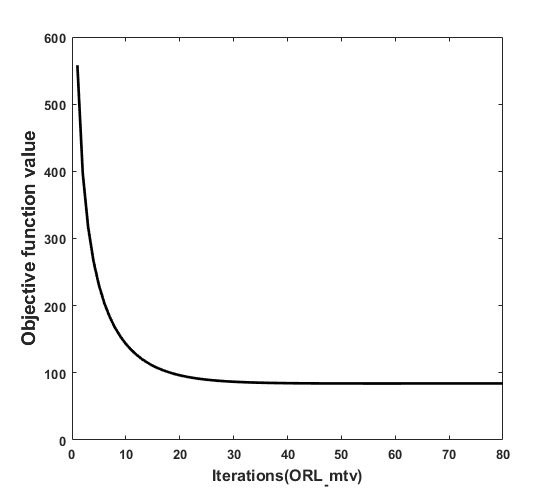}
\end{minipage}%
}%
\subfigure[]{
\begin{minipage}[t]{0.5\linewidth}
\centering
\includegraphics[width=1.9in]{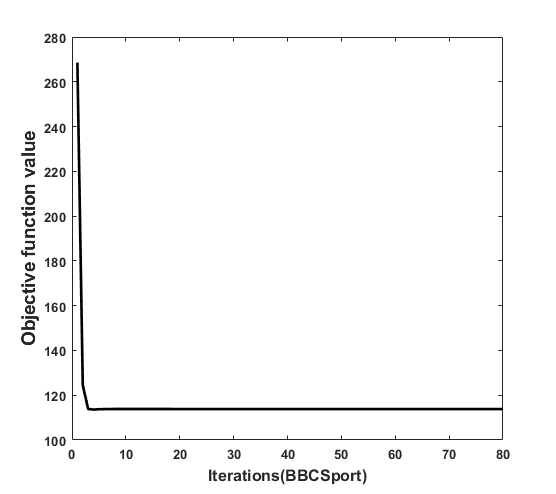}
\end{minipage}%
}%
       
\subfigure[]{
\begin{minipage}[t]{0.5\linewidth}
\centering
\includegraphics[width=1.9in]{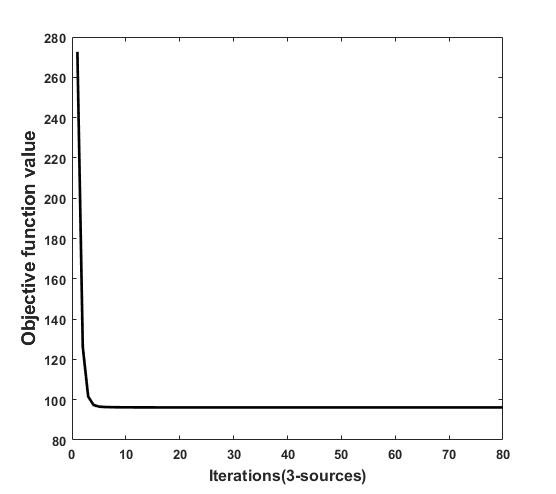}
\end{minipage}%
}%
\subfigure[]{
\begin{minipage}[t]{0.5\linewidth}
\centering
\includegraphics[width=1.9in]{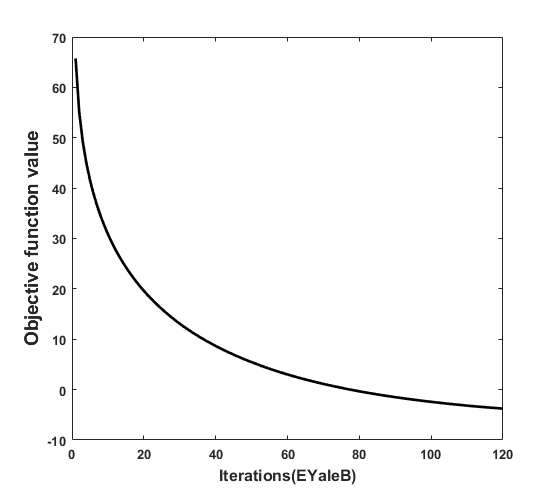}
\end{minipage}%
}%

\subfigure[]{
\begin{minipage}[t]{0.6\linewidth}
\centering
\includegraphics[width=1.9in]{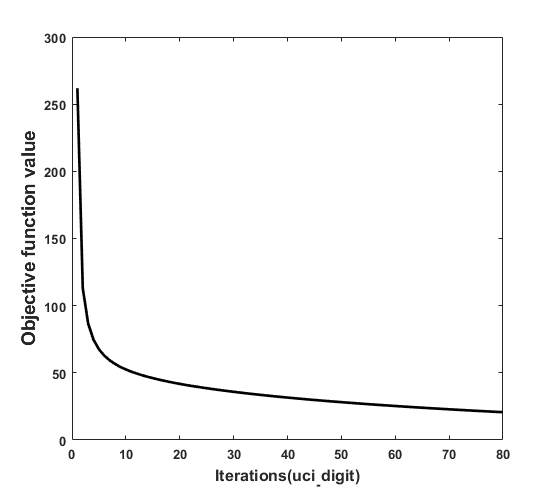}
\end{minipage}%
}%

\centering
\caption{Convergence curves for different datasets: (a) ORL\underline{\hspace{0.5em}}mtv; (b) BBCSport; (c) 3-sources; (d) EYaleB; (e) uci\underline{\hspace{0.5em}}digit. }
\label{figure7:convergence}
\end{figure}

\subsection{Clustering Results}
The experimental results of the different algorithms are shown in Tables \ref{tb2:ORL}-\ref{tb6:uci} for the ORL\underline{\hspace{0.5em}}mtv, 3-sources, BBCSport, EYaleB and uci\underline{\hspace{0.5em}}digit databases, respectively. Tables \ref{tb2:ORL}-\ref{tb6:uci} clearly show that our proposed algorithm achieves very promising results for these databases. A detailed analysis is conducted as follows:

Table \ref{tb2:ORL} shows the experimental results for the ORL\underline{\hspace{0.5em}}mtv dataset, which provides perfect clustering performance. Table \ref{tb2:ORL} shows that our algorithm achieves significant improvements of 6\%, 1\%, 2\%, 7\%, 3\%, 5\% and 6\% over the second-best result in the indexes of ACC, NMI, purity, precision, recall, F-score and AR, respectively. In addition, MLSSC has poor performance. This is because MLSSC uses the traditional nuclear norm and $L_1$-norm to realize low-rank and sparse representation of the similarity matrices. However, we have solved the shortcomings of traditional low-rank and sparse representation to a certain extent by using novel low-rank and sparse representation constraints. Moreover, MLSSC obtains the diversity between different views by means of mutual differences between views. This only guarantees that the obtained view matrix fuses some common information of each view but does not make full use of the specific information contained in each view.

Table \ref{tb3:3-sources} shows the experimental results for the 3-sources dataset, which is a text dataset. The classification results for the text dataset are not as stable as those for the image dataset since the text dataset contains fewer feature points and has a tight context. Strong semantic information is required to perform inter-class and intra-class discrimination. Therefore, the final classification results fluctuate greatly. However, our algorithm again exhibits great improvement, while some comparison algorithms achieve poor performance, such as MVGL and ECRMSC. The main reasons can be summarized as follows: 1) MVGL is adaptive graph learning; it only considers the retention of the local structure of the similarity factor of the original data matrix and the cooperative representation of the similarity matrix of each view but does not fully consider the information redundancy and other information. 2) ECRMSC uses the $L_1$-norm as a constraint to extract the complementary information between different sparse subspace matrices. However, ECRMSC does not make full use of the common information of each view and the retention of local structural features in the sparse representation process. As a comparison, our proposed algorithm has been improved on this basis, and the results are greatly improved (at least 3\%) for some evaluation metrics.

Table \ref{tb4:BBCSport} displays the clustering performance for the BBCSport dataset. This is a test dataset, and the final classification results fluctuate greatly. Table \ref{tb4:BBCSport} shows the effect of LMSC on the synthesis, which is only the last third, when our proposed algorithm achieves the best performance. Although LMSC exhibits a certain gap in performance with that of our proposed algorithm, there are certain similarities in the algorithm models. For example, both algorithms decompose the original matrix into sparse representation and noise error parts. Additionally, low-rank constraints are imposed on the sparse representation matrix and the noise error matrix. From this, we can conclude that the low-rank processing we imposed on the sparse representation matrix and the noise error matrix is very helpful for improving the clustering effect.

	Table \ref{tb5:EYaleB} shows the experimental results for the EYaleB face dataset. Since this dataset and the ORL\underline{\hspace{0.5em}}mtv dataset are both face datasets, the experimental results for these two datasets are similar. First, their clustering effects are better and shows a greater improvement. Second, the results for the two datasets that have been clustered multiple times are very close for each indicator. The only difference is that the EYaleB dataset and other comparison algorithms exhibit polarization in the clustering results. Some algorithms do not perform very well for this dataset, such as DiMSC and MVGL. Since one sample of each view in the EYaleB dataset has more than 10,000 features, these features may contain some redundant information causing noise, and the clustering performance will be greatly affected. In this way, we need to perform sparse representation and dimensionality reduction on the dataset. However, DiMSC and MVGL do not consider the noise matrix or remove redundant information, which decreases their performance.
	
The experimental results for the uci\underline{\hspace{0.5em}}digit dataset are shown in Table \ref{tb6:uci}. The performance of the proposed method for this dataset is relatively stable compared to that of the other algorithms. The experimental results of our proposed algorithm show slight improvement compared to the second-best results in some evaluations, such as improvements of 0.13\%, 0.45\% and 0.47\% for NMI, purity, and recall in the evaluation metrics, respectively. However, there is a remarkable improvement over the second-best comparison algorithm for some evaluation metrics, such as ACC, precision, F-score and AR. The improvements of the experimental results for these metrics are approximately 2\%, 3\%, 3\% and 2\%, respectively. 
\subsection{Similarity Matrix Analysis}

\begin{figure}[]
\vspace{-0.cm} 
\setlength{\abovecaptionskip}{-0.0cm}  
\setlength{\belowcaptionskip}{-0.1cm}  
\centering

\subfigure[ORL\underline{\hspace{0.5em}}mtv]{
\begin{minipage}[t]{0.3\linewidth}
\centering
\includegraphics[width=1.0in]{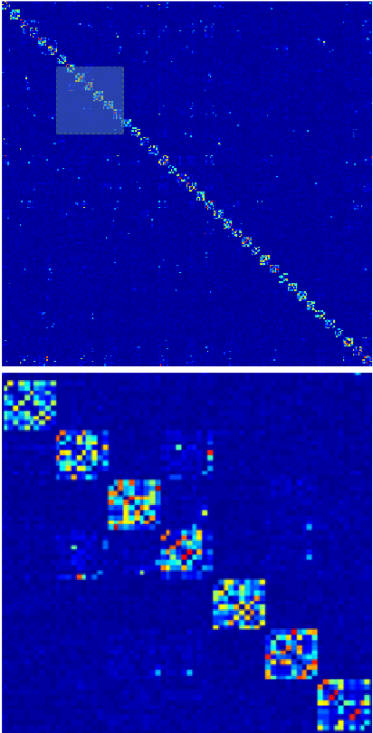}
\end{minipage}%
}%
\subfigure[EYaleB]{
\begin{minipage}[t]{0.3\linewidth}
\centering
\includegraphics[width=1.0in]{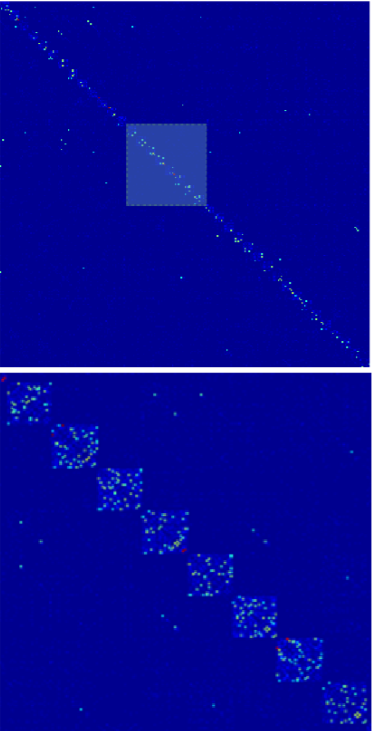}
\end{minipage}%
}%
\subfigure[uci\underline{\hspace{0.5em}}digit]{
\begin{minipage}[t]{0.3\linewidth}
\centering
\includegraphics[width=1.0in]{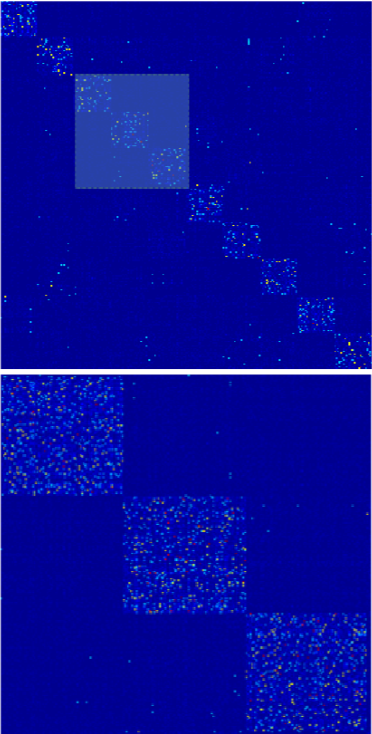}
\end{minipage}
}

\centering
\caption{The global similarity matrix A obtained by Eq. (33).}
\label{fig6:The global similarity}
\end{figure}

In this sub-section, the global similarity matrix is analysed. The similarity matrix can interpret the clustering performance effect very well. Because of space limitations, the global similarity matrices of only three datasets, the ORL\underline{\hspace{0.5em}}mtv, EYaleB and uci\underline{\hspace{0.5em}}digit datasets, are shown in Fig. \ref{fig6:The global similarity}. The first line in Fig. \ref{fig6:The global similarity} represents the original generated visual pictures. To show them clearly, we partially zoom in, and the shaded area in the first line in the figure is enlarged. The second line shows the visualization of the shaded area in the first line. From the second line in Fig. \ref{fig6:The global similarity}, we can see that different classes are clearly divided into squares. The more concentrated on the diagonal lines and the smoother it is outside the boxes, the better the clustering effect is. It can be seen that all three results are very good.

\subsection{Convergence Analysis}
To solve Eq. (10), we update each variable in the form of a locally optimal solution. For an optimization model, the convergence of the objective function is very important. Therefore, to intuitively explain the convergence of the proposed model, we present the analysis in Fig. \ref{figure7:convergence}. It shows that the BBCSport and 3-sources datasets converge very quickly and need only 3 iterations to converge, as they are two text datasets. In addition, the EYaleB dataset reports such a large number of iterations. According to the complexity analysis of the algorithm, we propose that the computational effectiveness of the method depends only on the number of clustering categories, the number of samples and the dimension of samples. First of all, the dataset EYaleB is a gray scale face dataset and has 1102 samples, which contains the faces of 38 different individuals. Second, the dimensions of each sample are very large containing 32,256 features. This is very complex. Therefore, when this data set is applied to our proposed algorithm, its iteration times are much more than that of other data sets. By comparison with the algorithm complexity in this paper, such as MLSSC$(O(tn_v n^3))$, MVGL$(tn_v^2 n^2)$, LRPP-GRR$(O(t(n^3+cn^2)))$, L0-LSSC$(O(tn^3+mn^2 ))$.The complexity of the proposed algorithm is reasonable in theory.

\section{Conclusion}
In this paper, we propose a new multi-view clustering algorithm for a low-rank sparse subspace. We use two decomposition methods to decompose the original data matrix. One method decomposes the original data into global sparse subspace and multi-view error matrices, and the other method decomposes the original data into some multi-view sparse subspace matrices. The two parts are combined by a regularization norm, resulting in an optimal global matrix and a greater diversity of individual view features. Furthermore, we use a new low-rank and sparse norm constraint for the defects of the nuclear norm and $L_1$-norm. The similarity matrix of the proposed method is adaptive. Finally, we put these sub-modules in a unified optimization framework and use the ADMM for optimization updates. Experiments are carried out for five well-known public datasets to verify the effectiveness of the proposed algorithm compared with several similar state-of-the-art algorithms. The experimental results show that the proposed method obtained the best results.

\ifCLASSOPTIONcaptionsoff
  \newpage
\fi

\bibliographystyle{IEEEtran}
\bibliography{egbib}




\end{document}